# Siamese Network with Dual Attention for EEG-Driven Social Learning: Bridging the Human-Robot Gap in Long-Tail Autonomous Driving

Xiaoshan Zhou[1], Carol C. Menassa[2], and Vineet R. Kamat[3]

[1]Ph.D. Candidate, Dept. of Civil and Environmental Engineering, University of Michigan, Ann Arbor, MI, 48109-2125. Email: xszhou@umich.edu

[2]Professor, Dept. of Civil and Environmental Engineering, University of Michigan, Ann Arbor, MI, 48109-2125. Email: menassa@umich.edu

[3]Professor, Dept. of Civil and Environmental Engineering, University of Michigan, Ann Arbor, MI, 48109-2125. Email: vkamat@umich.edu

**Abstract**

Robots with wheeled, quadrupedal, or humanoid forms are increasingly far-reaching in everyday life. However, unlike human social learning, they lack a critical pathway for intrinsic cognitive development, i.e., learning from human feedback, especially in real-time interactions. Human ubiquitous observation, supervision and shared control can offer robots valuable information (e.g., preferences, behavior, etc.). However, we need the ability to make robots understand the human's thoughts—mind reading—to truly align with them. Existing brain-computer interfaces (BCIs) are typically developed for narrow and well-defined tasks—such as motor imagery, emotion recognition, or sleep staging—based on public datasets. However, to evolve BCIs into scalable, real-world tools that support the crowdsourcing of human feedback, new approaches to user calibration and signal classification algorithms are urgently needed. This paper aims to shed light on two critical questions. First, to what extent can neural networks capture human thoughts as represented in sensory information encoded in scalp-recorded EEG signals? Second, what form of cognitive architecture could enable a BCI agent to categorize different mental processes—particularly when incorporating mechanisms of attention and memory? As a timely and motivating co-robotic engineering application, this work presents how BCIs can be leveraged to identify edge cases associated with the perennial long-tail problem in autonomous driving. We develop a mental imagery-based onboarding paradigm for BCI users, and design a Siamese convolutional network with dual-attention mechanism—Squeeze-Excitation for feature-wise attention and Multi-Head Self-Attention for temporal dynamics attention—for signal classification in a few-shot learning setting. Dynamic Time Warping Barycenter Averaging is introduced to derive representative signal templates for similarity comparison while accounting for temporal shifts across trials. Results from source localization indicate user engagement in generating neural signals spanning perception–action regions. Furthermore, the classification model achieves nearly a 100% increase in the integrated gradient measure, suggesting enhanced utilization of influential features compared to the baseline state-of-the-art approach. We further discuss how the model supports inter- and intra-subject adaptation. This research lays the groundwork for scalable human-in-the-loop annotation to address the long-tail problem in autonomous driving and presents an adaptable approach to enable socially guided learning in other robotic systems.

**Keywords:** Long-tail problem, Electroencephalogram (EEG), Human feedback, Siamese network, Multi-head self-attention mechanism, Source localization, Dynamic Time Warping (DTW), Autonomous vehicles, Robotic driving

## 1 Introduction

The human brain has evolved over thousands of years to become the most intelligent biological system in nature (Dehaene, 2020). Compared to other primates, the human frontal lobes are not only larger but also more densely populated with neurons and exhibit more complex dendritic structures and interconnections (Smaers et al., 2017). Since the $17^{th}$ century, researchers have investigated the brain's anatomical structure and sought to replicate neural functions in artificial systems (Macpherson et al., 2021). One notable example is the auto-associative network, which learns through the local co-activation of presynaptic and postsynaptic firing (Treves & and Rolls, 1991). The most successful modern neural networks rely on backpropagation to update weights, which, however, lacks biological plausibility (Rolls, 2023), as human neurons do not communicate across such long distances with centralized error signals (Plebe & Grasso, 2019). Despite this disconnect, neural networks have achieved remarkable success in processing both temporal (e.g., sequential EEG data) and spatial (e.g., image-based) information. However, there remains an ongoing debate over whether these models truly "understand" their inputs or merely exploit statistical regularities. Brain–computer interfaces (BCIs) represent a compelling intersection between neuroscience and machine learning, and the field is undergoing rapid expansion—from traditional applications in rehabilitation and clinical settings (Dobkin, 2007; Grosse-Wentrup et al., 2011) to emerging domains such as vehicles (Zhang et al., 2023), assistive robotics (Naser & Bhattacharya, 2023), and consumer-facing technologies (Ramsey & Vansteensel, 2025).

BCI systems rely heavily on machine learning techniques to decode neurophysiological signals. Convolutional neural networks (CNNs), in particular, have emerged as state-of-the-art (SOTA) tools for mapping EEG patterns to user intentions (Li et al., 2025; Tabar & Halici, 2017). In recent years, Siamese convolutional networks have demonstrated exceptional performance in few-shot learning scenarios, enabling robust classification with minimal training data—a critical factor in reducing onboarding time for new BCI users (Ahuja & Sethia, 2024). These models consist of multiple hidden layers that perform weighted summations, nonlinear transformations, and feature extraction operations. Nevertheless, despite their success in accuracy metrics, prior studies often overlook the question of whether these models genuinely capture the most relevant neural features. Without interpretability, a model trained on one individual and one task may fail to generalize to others. Although methods like transfer learning and continual learning (e.g., experience replay) offer partial solutions to domain adaptation (Zhou & Liao, 2023b), the next major challenge is to move beyond isolated technical fixes. There is a pressing need to develop generalizable BCI frameworks capable of adapting across individuals, tasks (e.g., motor control, speech decoding, attention monitoring), and domains (e.g., medical, gaming, assistive technology).

**1.1 Research Questions and Motivating Case Study**

In this context, this paper aims to shed light on four critical questions. First, can neural networks really "understand" unstructured Electroencephalogram (EEG) signals? Most existing machine learning models are trained on clean, public EEG datasets based on well-defined tasks (like imagining left vs. right hand movement). But real life involves intricate situations, with more complex connections between perception and action. Furthermore, EEG signals recorded from the scalp are subject to attenuation and spatial smearing due to volume conduction, raising an important concern: when these signals are fed into neural networks, and certain features emerge as dominant in predicting mental states, can we confidently assert that these features reflect meaningful neural activity? Or are we overfitting to statistical artifacts that do not represent underlying neuronal processes?

Second, if we accept that scalp EEG signals do indeed encode perceptible and differentiable mental states—even for more complex or vague tasks—then it becomes equally important to consider the user's ability to reliably generate those states. In other words, improving machine learning models is only part of the solution; training users to consistently control their mental states may be just as crucial. To support this goal, we propose the development of a structured BCI onboarding protocol that guides users to produce stable and interpretable brain activity, and we gathered experimental data to show that this is feasible in a preliminary proof-of-concept form.

Third, the neural network architecture for signal decoding needs to be able to generalize across individuals and tasks. EEG signals are high-dimensional, so selecting meaningful features before feeding them into the model is critical. Some tasks have well-established neural markers—such as contralateral event-related desynchronization (ERD) during motor imagery (Formaggio et al., 2010), P300 potentials elicited around 300ms in oddball paradigms (Li et al., 2019), and error-related potentials following conflicting responses (Navarro Cebrian et al., 2013). But for less structured tasks, there are no clearly defined features, and even within the same task, the identified features could be inconsistent: Different users may think about things differently (e.g., visual vs. kinesthetic imagery) (Neuper et al., 2005); personal traits (e.g., risk-taking vs. cautious behavior) can affect how the brain responds (Zhou et al., 2022). Moreover, several cognitive functionality—such as mediating actions—lack clear neural correlates.

Both prospective neural activity (Doll et al., 2015), such as readiness potential preceding voluntary movement (Colebatch, 2007), and post-action EEG fluctuations (Kornhuber & Deecke, 2016) signifying reevaluation (Gershman et al., 2014) have been observed, but their patterns remain inconsistent across contexts. Given these uncertainties, the model architecture must be able to automatically focus on relevant spatial and temporal parts of the EEG signal, while also accommodating temporal misalignments and variations within trials of the same class, especially when there's limited training data.

Finally, in similarity-based learning (e.g., using templates), the ability to store and retrieve reference signals efficiently is crucial. As BCI systems transition into real-world applications, it is imperative to start establishing a data infrastructure that can scale, support fast, reliable template recall, and remain adaptable to new users and new data over time.

To illuminate the core questions raised in this work, we turn to a representative use case in autonomous driving. A key challenge in autonomous driving is the long-tail problem: rare, unexpected events (like unusual pedestrian behavior (McCausland, 2019) or sensor glitches (Wessling, 2023)) that are difficult to capture in typical training datasets (Liu & Feng, 2024). Although some researchers try to address this data scarcity by statistical methods (e.g., better sampling (Ding et al., 2023) or simulation (Chen et al., 2020; Zhao et al., 2024)), we argue that an underutilized but more efficient strategy is to leverage human feedback as a natural filter to identify and highlight significant moments of interest. "Feedback" in our definition includes what drivers observe and how they react in real time. Right now, we are in the phase of partial driving automation (Phase II, as defined by (SAE., 2018)), where humans are required to play an active supervisory role in today's semi-autonomous driving, and this monitoring process naturally produces rich human insights. The drivers are constantly observing, but their cognitive responses—gut reactions, surprise, hesitation—are difficult to capture through language or post-hoc reporting.

Therefore, we propose using real-time brain signals (like EEG) as a window into human cognition: brain activity becomes a filtering mechanism, surfacing events that merit further investigation for data curation, system debugging, and even learning. We see this approach as having strong potential to scale into a crowdsourcing initiative. Accordingly, the focus of this paper is to provide the necessary infrastructure and algorithmic support for such an EEG-based BCI system. Once established, this framework could be extended to identify a wide range of infrastructure flaws and latent environmental risks. In addition, the collected EEG signals also offer a rich source of human cognitive intelligence, which could support future efforts in rule-based expert systems, representation learning, and imitation learning. These directions lie beyond the scope of the current paper, but they are explored in prior work by the authors (Zhou & Liao, 2023a).

## 2 Knowledge Gaps and Research Contributions

### 2.1 Siamese Network-based Few-Shot Learning for EEG Signal Classification

Using CNN to classify EEG signals, based on a bibliometric search, there are over 4000 published papers. This is because there are many public EEG datasets for researchers to develop and test models. But when it comes to few-shot learning for EEG, i.e., learning and classifying EEG signals with a very limited number of trials, the number of studies drastically reduces to approximately three dozen. The Siamese network is dominantly used given its similarity-learning nature and prior success in visual image classification under few-shot settings (Chicco, 2021). For example, Munia et al. (2021) developed a Siamese CNN to categorize EEG signals for epileptic seizure detection. This is particularly useful because the available data from patients are limited and imbalanced. They found an 11.66% increase in F1-measure compared with a baseline CNN, and thus claimed that the model is capable of capturing distinguishable morphological characteristics in the EEG signals. Similarly, (Calhas et al., 2020) employed a Siamese neural network for schizophrenia diagnosis based on EEG data from limited clinical trials, and achieved SOTA performance.

(You et al., 2023) developed a Siamese network to generate latent features from EEG epochs and use similarity comparisons to categorize different sleep stages. They used single-channel EEG data and tested on two public sleep staging datasets, demonstrating superior performance than SOTA. To leverage multi-channel signal decoding, (Shen et al., 2022) incorporates a cross-channel fusion module into a Siamese CNN for motor imagery (MI) classification. Their model was tested on public BCI Competition datasets and outperformed other SOTA MI EEG signal classification methods. To further leverage brain connectivity in addition to localized activation, (Zeng et al., 2022) developed a Siamese Graph Convolutional Network for emotion recognition. Instead of stacking graph convolution layers, they implemented a deep attention layer to extract valuable EEG features, and achieved superior performance on two public emotional EEG datasets compared to baseline methods. Beyond diagnostic systems, Siamese networks were among the earliest models used for EEG-based identity verification that leverages biometric traits to replace signatures, which are vulnerable to forgery. (Das Chakladar et al., 2021) modified the Siamese network architecture by integrating two encoders to leverage multimodal biological data sources, and achieved SOTA performance in biometric authentication.

In addition, (Fu et al., 2022) leveraged a Siamese Neural Network to learn highly similar EEG features across different subjects. This approach was proposed as a data augmentation technique and demonstrated better performance than common classifiers. Similarly, (Zhang et al., 2022) applied a Siamese CNN to learn representations from EEG signals, reducing the complexity of the signal into a low-dimensional embedding. They extracted distance-based representations from pairwise EEG data and combined this feature extractor with a k-Nearest Neighbor classifier. This model was applied to limb neurorehabilitation datasets collected from stroke patients to recognize their motor intentions. This was the first application of a Siamese CNN as a feature extractor for non-stationary EEG signals in BCI-assisted post-stroke rehabilitation, and demonstrated superior performance compared to existing methods. Similarly, the Siamese neural network has been used for discriminative features learning from EEG signals for imagined speech classification as a potential intuitive communication tool for BCI (Lee et al., 2020),

These studies demonstrated that Siamese networks consistently outperform SOTA models in few-shot EEG classification, and that convolutional layers in CNN are effective at extracting meaningful EEG features. Building upon these findings, we propose to leverage the Siamese similarity-learning architecture as the backbone for representation learning (compared to conventional supervised learning), and embed convolutional layers to extract spatiotemporal patterns from EEG data inputs. However, current Siamese architectures are designed for pre-known tasks and do not consider domain adaptation (e.g., adapting to new users, datasets, or tasks). They do not explicitly address inter-subject (between individuals) and intra-subject (within the same individual over time) variability. Thus, our proposed modifications are focused on allowing the model to extract relevant features from EEG data while dynamically adjusting which features to prioritize.

**2.2 Attentional Mechanism in Deep Neural Network for EEG Signal Classification**

The integration of attention mechanisms into EEG-based deep learning architectures has recently gained momentum due to their potential to improve classification performance across a range of tasks. (Eldele et al., 2021) used multi-head attention to help a CNN model capture how EEG signals change over time, for classifying sleep stages using signals from just one EEG channel. They called this "adaptive feature recalibration". Similarly, (Tao et al., 2023) proposed an attention-based convolutional recurrent neural network to extract more discriminative features for emotion recognition. They use channel-wise attention to adaptively weigh different EEG channels and self-attention to identify which time steps in the EEG sequence are more important. (Song et al., 2023) applied self-attention to extract global correlations within local temporal features (learned by CNN), and tested their model on MI and emotion recognition datasets—achieving SOTA performance. (Altaheri et al., 2023) also applied multi-head self-attention to focus on the most informative features in MI EEG, and combined these with temporal features extracted by CNNs. Their model also outperformed SOTA benchmarks on public datasets.

These recent studies find attention mechanisms for EEG classification to be promising because they improve model focus on relevant features and increase accuracy. However, most current approaches add attention modules as external layers or plug-ins to CNNs. They are not deeply integrated or biologically motivated. Moving forward, we advocate for a more principled approach: drawing inspiration from how attention works in the human brain and using that understanding to design more effective, embedded attention modules for EEG deep learning.

Below we provide a brief review of the biological basis and functions of the human attention system, and how this understanding informs the development of artificial attention mechanism in machine learning. Three biological attention systems—alerting, orienting, and executive attention—provide a foundation for how the brain selects, amplifies, and prioritizes information.

First, the attention system keeps us alert and ready to respond to threats or urgent situations. When needed, it rapidly releases neuromodulators such as serotonin, acetylcholine, and dopamine, which travel through long-range axonal pathways to modulate activity across widespread cortical regions. Some researchers describe this as a "now print" signal, a mechanism by which the brain flags salient events for consolidation into long-term memory (Kastner, 2000).

Second, attention addresses the fundamental challenge of information overload. The brain continuously receives massive streams of sensory input, far exceeding its capacity for deep processing. To manage this, a hierarchical attention system selectively allocates cognitive resources to inputs that align with current goals. This selective processing is essential for learning; without attention, critical information would be lost amid irrelevant noise. In contrast, conventional artificial neural networks typically process all inputs equally, which is way less efficient. In 2014, Yoshua Bengio and Kyunghyun Cho introduced the first attention mechanism in deep learning for machine translation (Bahdanau et al., 2014). Their model dynamically focused on the most relevant parts of the input sequence, which significantly improved both learning speed and accuracy. Although scientists still debate whether attention is like a filter (ignoring irrelevant inputs) or an amplifier (boosting relevant ones) (Fazekas & Nanay, 2021), it is clear that it increases the activity of neurons related to the attended information while suppressing irrelevant signals. This differential modulation increases sensitivity to meaningful input and reorients downstream neural circuits to further reinforce the representation, aiding memory consolidation and decision-making (Kastner & Ungerleider, 2000).

Finally, attention underpins what psychologists refer to as the central executive—a set of control mechanisms that guide deliberate behavior, including sustained concentration and goal-directed action. Often equated with executive attention or self-control, this system enables the brain to maintain focus, inhibit distractions, and pursue a chosen course of action (Friedman & Robbins, 2022).

The three biological attention systems—alerting, orienting, and executive attention—provide a foundation for how the brain selects, amplifies, and prioritizes information. Inspired by these mechanisms, we integrate Excitation-Squeeze Networks and Multi-Head Self-Attention into deep learning architectures to replicate these cognitive functions in artificial neural networks.

The alerting attention system, responsible for global awareness and vigilance, aligns closely with the squeeze mechanism in Squeeze-and-Excitation Networks (SENets) (Hu et al., 2018). Just as the brain monitors broad sensory input before filtering relevant details, the squeeze step aggregates global information across feature maps in CNNs. This ensures that key patterns are retained while irrelevant details are compressed, improving the efficiency of feature selection. In computer vision tasks, SENets have been widely used in ResNet and EfficientNet, where the excitation step dynamically recalibrates feature importance, much like how the brain amplifies important neural signals in response to alerting stimuli.

The orienting attention system, which directs focus to specific sensory input and suppresses distractions, finds its counterpart in local self-attention mechanisms such as positional encoding and windowed attention in Transformers. In this context, self-attention restricts processing to local regions, mimicking the brain's ability to prioritize stimuli within a given spatial or temporal frame. This concept is fundamental in models like Vision Transformers, where patches of an image are analyzed with localized attention first, before aggregating global context (Dosovitskiy et al., 2020). Similarly, in EEG-based BCI applications, self-attention can be employed to focus on relevant neural patterns over time while ignoring transient noise (Song et al., 2023). Furthermore, multi-head self-attention, powerful in natural language processing models, including Transformers, help attend to different linguistic dependencies to improve contextual understanding (Vaswani et al., 2017). In EEG classification, multi-head self-attention allows the network to process multiple frequency bands or sensor channels concurrently, thus potentially enhancing adaptive prioritization and classification accuracy.

**2.3 Neural plasticity and Its Implications for BCI Onboarding**

BCI onboarding—helping new users learn how to use a brain-computer interface—is grounded in the principle of neural plasticity, i.e., the brain's inherent ability to reorganize its structure and function through learning and experience (Rossini et al., 2003). This adaptive capacity of neural circuits is central to BCI training. Compelling evidence from neurorehabilitation research demonstrates that users can recall motor skills to reshape neural representations, thereby improving prosthetic control. Over time, these adaptations give rise to stable cortical maps that can be readily reactivated for neuroprosthetic operation, resembling what some researchers describe as putative memory engrams (Ganguly & Carmena, 2009).

Studies on gait rehabilitation post-stroke (Belda-Lois et al., 2011; Takeuchi & Izumi, 2013) emphasize the role of neurophysiological interventions in promoting plasticity-driven recovery. Targeted training programs can induce dendritic growth, synaptic remodeling, and increased neurochemical activity at the cellular level. At the cortical level, such training can lead to structural and functional reorganization in regions such as the motor cortex and cerebellum (Arya et al., 2011). From a pathophysiological standpoint, plasticity manifests at multiple scales: microscopically through synaptic efficacy changes, latent pathway activation, and neurogenesis; and macroscopically through phenomena such as diaschisis, sensory substitution, and morphological reconfiguration. These mechanisms form the basis for therapeutic strategies—including pharmacological treatments, transcranial magnetic stimulation, neurosurgical approaches, and BCIs—that support motor and cognitive rehabilitation (Duffau, 2006; Hatsopoulos & Donoghue, 2009).

Although neural plasticity has traditionally been associated with long-term structural changes, for instance, the hippocampal adaption in taxi drivers due to spatial navigation-related demands (Maguire et al., 2000), recent research indicates that BCI training can induce plasticity within minutes or hours. For example, one hour of MI training in BCI-naïve participants has been shown to enhance functional connectivity and task-relevant brain activity (Nierhaus et al., 2021). Even 30 minutes of BCI-based neurofeedback training targeting kinesthetic MI of right-hand movements resulted in structural plasticity changes within interhemispheric sensorimotor networks (Kodama et al., 2023). These findings provide a foundation for developing an onboarding paradigm that helps BCI users generate consistent and classifiable neural patterns within a limited number of training trials. Assuming that even short-term engagement can yield repeatable neurophysiological signatures that can subsequently be captured by machine learning models, this opens up scalable pathways for personalized, rapid, and minimally burdensome BCI calibration.

Notably, BCI training heavily relies on covert, mental actions—where people modulate cognitive processes without overt physical movement. The emerging field of computational psychiatry investigates how precision-weighted neural processes govern these mental actions, offering insights into their neurophysiological underpinnings (Lecaignard et al., 2020). These concepts closely with work in contemplative neuroscience, where cognitive opacity and meta-awareness—such as those cultivated in meditation—are shown to enhance attentional control and self-regulation (Eberth & Sedlmeier, 2012; Tang et al., 2015). These are similar to what BCI users do for onboarding—learning to control brain activity through mental strategies. The framework of neurophenomenology (Lutz, 2002; Varela, 1996) also offers a valuable lens through which to study the subjective experience of BCI adaption. This is useful in BCI because it helps understand how users become aware of and learn to modulate their brain activity.

To summarize, this section provides a comprehensive review of current Siamese neural network architectures developed for few-shot EEG signal classification. It also highlights the role of attentional mechanisms in the human brain and how these insights inform the design of modern attention mechanisms in neural networks, which can be tailored specifically for EEG-based tasks. We further review the concepts of neural plasticity and supporting neuroanatomical evidence, and discuss how mental imagery can be leveraged to generate reliable and consistent mental activity—even for naïve BCI users. In the next section, we present the algorithmic design of our proposed BCI system, which combines similarity-based few-shot classification in a Siamese network framework with dual-attention mechanisms to improve robustness and generalizability.

## 3 Technical Approach

### 3.1 Few-Shot Classification Overview – Similarity learning

We propose to learn a function $f(z, x)$ that compares a reference EEG sequence $z$ to a candidate sequence $x$ and returns a high similarity score if the two sequences exhibit similar brain activity patterns, and a low score otherwise. To classify new EEG signals, the function can be used to compare an input sequence to stored reference patterns, and then select the candidate with the maximum similarity. An illustration of the similarity comparison is shown in Figure 1.

Given their success in visual object tracking (Bertinetto et al., 2016), we adopt a convolutional Siamese network as the base backbone of the function $f$. Siamese networks apply an identical transformation $\phi$ to both EEG inputs from pairs and then compare their representations using another function $g$, following $f(z, x) = g(\phi(z), \phi(x))$. A contrastive loss function is then computed and used for backpropagation. This loss encourages the network to learn smaller distances for pairs of instances from the same category (positive pairs) and larger distances for pairs from different categories (negative pairs), thereby fine-tuning the network parameters to enhance its discriminative ability.

However, conventional Siamese networks may fail to capture the dynamic nature of EEG signals, where different time steps and frequency ranges contribute unequally to the classification task. To address this, we introduce a Dual-Attention Siamese Network that incorporates both channel attention (to highlight the most informative EEG features) and temporal attention (to emphasize critical moments in the EEG sequence).

In addition, to measure the distance between two sequences for selecting the most discriminative feature while accounting for potential temporal stretches across trials, we use the Dynamic Time Warping (DTW) technique. Its extension was also introduced to create the templates that serve as the reference signals. Details are provided in the following subsections.

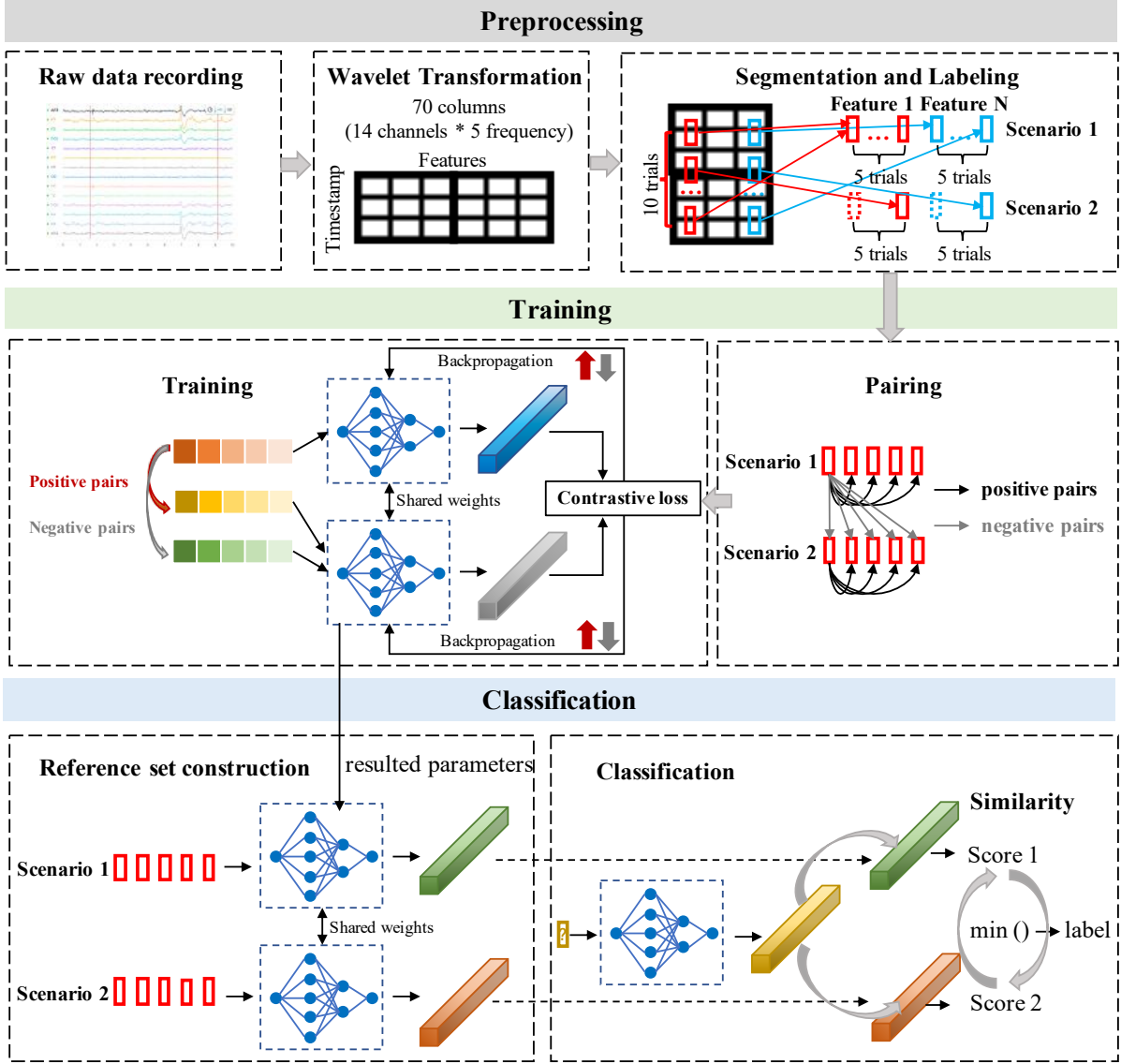

Figure 1. Step-by-step approach for constructing a Siamese network for EEG sequence classification.

### 3.2 Convolutional Siamese Architecture as Backbone

We define a function as fully-convolutional if it is invariant to time shifts in EEG sequences. Let $L_\tau$ denote the time-shift operator, where $(L_\tau x)[t] = x[t - \tau]$. A function $h$ that maps EEG signals to feature embeddings is said to be shift-invariant with integer stride $k$ if:

$$h(L_{k\tau}x) = L_\tau h(x) \tag{1}$$

for any temporal shift $\tau$. This property ensures that our model captures EEG features independently of when they occur in time.

### 3.3 Dual Attention Mechanism

Instead of processing individual EEG time points separately, we employ a convolutional embedding function $\phi$ that extracts hierarchical representations from entire EEG sequences. In addition, this function learns spatially-invariant (electrode-level) and temporally-invariant (time-shifted) features using a dual-attention mechanism.

#### 3.3.1 Channel Attention (Squeeze-Excitation)

The channel attention module learns to emphasize the most important EEG feature channels (e.g., electrodes and power bands) using a squeeze-excitation mechanism. We first use a global average pooling to summarize channel importance:

$$c_i = \frac{1}{T}\sum_{t=1}^{T} X_{t,i} \tag{2}$$

where $X \in \mathbb{R}^{T \times C}$ is the input EEG feature map. $T$ is the sequence length (time steps). $C$ is the number of feature channels. $c_i$ is the global average-pooled feature for channel $i$. Fully connected layers are introduced to learn attention weights:

$$a_c = \sigma(W_2 \cdot ReLU(W_1 c + b_1) + b_2) \tag{3}$$

where $W_1 \in \mathbb{R}^{C \times d}$, $W_2 \in \mathbb{R}^{C \times d}$ are the weights we try to train. $b_1$ and $b_2$ are introduced as biases. $\sigma$ is the sigmoid activation function to normalize attention scores. The output $a_c$ is the attention score vector for each EEG feature channel. Finally, the attention weights are applied to the input feature map via element-wise multiplication:

$$X'_{t,i} = a_{c,i} \cdot X_{t,i} \tag{4}$$

such that $X'$ is obtained as the reweighted EEG feature map after channel attention.

#### 3.3.2 Temporal Attention (Multi-Head Self-Attention)

The temporal attention modules learn to assign importance to different time steps using a multi-head self-attention. To compute the temporal importance, each time step $X_t$ is first transformed into query, key, and value representations:

$$Q = XW_Q, K = XW_K, V = XW_V \tag{5}$$

where $W_Q$, $W_K$, $W_V \in \mathbb{R}^{C \times d}$ are projection matrices we try to train. d is the feature dimensionality per head. For each head, the attention-weighed representation at time $t$ is computed based on

$$A_t = softmax(\frac{QK^T}{\sqrt{d}})V \tag{6}$$

The scaled dot product attention function aims to make more important time steps receive higher weights. The softmax function is used to ensure that the attention weights sum to 1 across all time steps. The attention outputs from all heads are concatenated and projected back to the original shape:

$$MHA(X) = W_O \cdot Concat(A_1, A_2, \ldots, A_H) \tag{7}$$

supposing that there are $H$ heads. $W_O$ is also a learnable projection matrix. Layer normalization is followed to stabilize training. Finally, the learned temporal attention weights are applied to the input feature map:

$$X''_{t,i} = MHA(X) \cdot X_{t,i} \tag{8}$$

where $X''$ is the reweighted EEG feature map after temporal attention.

### 3.3.3 Residual Connection

To prevent vanishing gradients and allow smoother gradient flow, we use residual connections instead of product to obtain the final EEG feature map:

$$X_{final} = X + X' + X''$$

$$\tag{9}$$

### 3.4 Dataset (Pairs) Construction

The EEG data was processed into a tabular structure with timestamps, EEG channel power values, and corresponding event markers. To give a more precise definition, let $X$ be the evoked EEG power dataset: $X = \lfloor (t_i, x_i) \rfloor_{i=1}^{N}$, where $t_i$ is the timestamp at sample $i$, $x_i \in \mathbb{R}^C$ is the EEG power vector across $X$ feature channels at time $t_i$. Events are marked by their onset times: $E = \lfloor (e_k) \rfloor_{k=1}^{K}$, where $e_k$ represents an event onset timestamp for event $k$.

Each epoch is extracted using a sliding window, where the window length is determined by the interest period $T_{interest}$ and the sampling rate $f_s$. The total number of samples in each window is:

$$L = T_{interest} \times f_s$$

$$\tag{10}$$

Thus, EEG epoch is extracted for an event $e_k$ as:

$$\mathcal{E}_k = \{X_j | t_j \in [e_k, e_{k+L}]\} \tag{11}$$

Subsequently, the extracted EEG epochs are paired based on event similarity.

Positive pairs ($y=1$) are defined as:

$$(X^A, X^B, y) = (\mathcal{E}_i, \mathcal{E}_j, 1), if\ e_i = e_j \tag{12}$$

Otherwise, negative pairs ($y=0$) are defined as:

$$(X^A, X^B, y) = (\mathcal{E}_i, \mathcal{E}_j, 0), if\ e_i \neq e_j \tag{13}$$

### 3.5 Training with EEG Sequences

We employ a discriminative approach, training the network on positive and negative EEG sequence pairs using a binary cross-entropy loss function:

$$L(y, v) = -y\log(v) - (1-y)\log(1-v) \tag{14}$$

where $v$ is the predicted similarity score between two EEG sequences. $y \in (0,1)$ is the ground truth label indicating whether the two sequences belong to the same class (1 for similar, 0 for different). We use a fully convolutional embedding function $\phi$ to extract shared feature representations for both inputs. The two transformed EEG feature maps are then compared using an element-wise absolute difference function $g$, yielding a similarity score:

$$f(z, x) = g(\phi(z), \phi(x)) = |\phi(z) - \phi(x)| \tag{15}$$

This ensures that the model learns to map similar EEG sequences closer together in feature space while keeping dissimilar sequences apart.

During training, the network optimizes the parameters $\theta$ using the Adam optimizer to minimize the loss function:

$$\theta^* = argmin_\theta \mathbb{E}_{(z,x,y)}[L(y, f(z, x; \theta))] \tag{16}$$

where the expectation is taken over all EEG sequence pairs in the dataset.

### 3.6 Dynamic Timing Warping for Feature Selection

Dynamic Time Warping (DTW) measures distance between two sequences while accounting for variations in timing and speed (Müller, 2007). Given two EEG sequences $X$ and $Y$, DTW finds an optimal alignment path that minimizes the cumulative distance between corresponding time points. The DTW distance is defined as:

$$DTW(X,Y) = min_\pi \sum_{(i,j)\in\pi} d(X_i, Y_j) \tag{17}$$

where $X, Y \in \mathbb{R}^{T \times C}$ are EEG sequences for length $T$ and $C$ feature channels, $d(X_i, Y_j)$ is the distance metric. $\pi$ is the warping path that aligns $X$ and $Y$ in time.

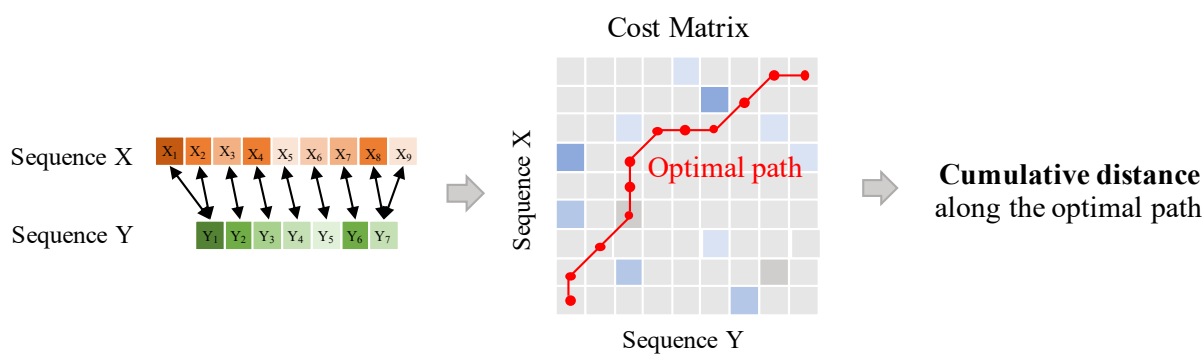

Figure 2: Illustration of the DTW process.

Figure 2. illustrates the steps for constructing DTW. The detailed procedure with pseudocode can be found in Figure 3.

**Algorithm 1** Pseudocode for Dynamic Time Warping (DTW)

**Input:** Two datasets:

$$\text{Dataset1} = \{T_1^{(1)}, T_2^{(1)}, \ldots, T_n^{(1)}\}, \quad \text{Dataset2} = \{T_1^{(2)}, T_2^{(2)}, \ldots, T_m^{(2)}\}$$

Each trial $T$ contains features:

$$\text{Features} = \{f_1, f_2, \ldots, f_{70}\}$$

**Output:** The feature $f_{\max}$ with the maxmium median distance.
1: Initialize the cost matrix $g(i, j)$:
2: $g(0, 0) = 0$
3: **for** $i = 1$ to $|T_1^{(1)}|$ **do**
4:   $g(i, 0) = \infty$
5: **end for**
6: **for** $j = 1$ to $|T_1^{(2)}|$ **do**
7:   $g(0, j) = \infty$
8: **end for**
9: **for** each feature $f_k \in$ Features **do**
10:  **for** each trial pair $T_i^{(1)} \in$ Dataset1, $T_j^{(2)} \in$ Dataset2 **do**
11:    **for** $i = 1$ to $|T_1^{(1)}|$ **do**
12:      **for** $j = 1$ to $|T_1^{(2)}|$ **do**
13:        Compute:

$$g(i, j) = \min \begin{cases} g(i-1, j-1) + d(T_{i,k}^{(1)}, T_{j,k}^{(2)}) \cdot w_D, \\ g(i-1, j) + d(T_{i,k}^{(1)}, T_{j,k}^{(2)}) \cdot w_V, \\ g(i, j-1) + d(T_{i,k}^{(1)}, T_{j,k}^{(2)}) \cdot w_H \end{cases}$$

14:      **end for**
15:    **end for**
16:    Trace back the optimal path from $g(|T_i^{(1)}|, |T_j^{(2)}|)$ to $g(0, 0)$.
17:    Compute the cumulative distance:

$$D_{k,i,j} = \sum_{(p,q) \in \text{OptimalPath}_{k,i,j}} d(T_{p,k}^{(1)}, T_{q,k}^{(2)})$$

18:    Append $D_{k,i,j}$ to CumulativeDistances$_k$.
19:  **end for**
20:  Compute the median distance:

$$\text{MedianDistance}_k = \text{median}(\text{CumulativeDistances}_k)$$

21: **end for**
22: Identify the feature with the maxmize median distance:

$$f_{\max} = \arg\max_k(\text{MedianDistance}_k)$$

23: **return** $f_{\max}$

Figure 3. Pseudocode for DTW.

## 3.7 Reference Set Construction

To create a reference representation for each event type, we use Dynamic Time Warping Barycenter Averaging (DBA) (Petitjean et al., 2011). This method computes a representative EEG template for each event by averaging multiple EEG epochs while aligning them in the time domain using DTW. Given a set of $M$ EEG sequences $[X_1, X_2, ..., X_M]$, the DBA reference set is:

$$B_k = argmin_B \sum_{m=1}^{M} DTW(B, X_m) \tag{18}$$

where $B_k$ is the barycentric template for event $e_k$. The optimal $B_k$ minimizes the sum of DTW distances across all epochs.

## 3.8 Implementation Details

The initial values of the parameters of the embedding function follow the Xavier (Glorot Uniform) initialization (Glorot & Bengio, 2010). Training is performed using mini-batches of size 32, and an exponential learning rate decay schedule is applied, where the initial learning rate $\eta_0 = 10^{-3}$ is annealed geometrically every 100,000 steps using a decay factor of 0.96, approaching $10^{-5}$ over time. Batch normalization is applied after each convolutional layer to stabilize training. To prevent overfitting, dropout (0.3) is applied in both convolutional and fully connected layers, and we apply an early stopping strategy, monitoring the validation loss with a patience threshold of 10 epochs. The best model weights, determined by the lowest validation loss, are stored for evaluation.

In this section, we introduced the technical framework for training the BCI agent to distinguish between different mental activities. The training process is based on similarity learning, where the system learns to compare new EEG inputs against representative templates of known mental states. Specifically, we use DBA to construct these templates from training data, while also training a Siamese CNN with Dual Attention that learns to measure the similarity between EEG sequences and the stored templates. Below, we will demonstrate how this framework works seamlessly with our proposed fast onboarding protocol. By combining both components, the system is able to quickly adapt to individual users while maintaining high classification accuracy, thereby improving the overall usability and personalization of the BCI system.

## 4 Research Methodology

### 4.1 Motivating Engineering Problem

Recent investigations into autonomous vehicle (AV) accidents have highlighted a persistent challenge in the field: the inability of current systems to effectively handle rare and unpredictable scenarios, often referred to as corner cases (Liu & Feng, 2024). These events—ranging from the failure to detect jaywalking pedestrians (McCausland, 2019) to misinterpreting the structure of complex vehicles such as articulated buses (Wessling, 2023)—indicate the limitations of data-driven and end-to-end perception and planning algorithms when confronted with out-of-distribution inputs. Although relatively infrequent, these scenarios are disproportionately responsible for severe failures and safety-critical outcomes, and raise fundamental questions about the robustness and generalization capacity of AV systems.

Despite advances in simulation (Chen et al., 2020; Ding et al., 2023; Hanselmann et al., 2022), synthetic environments fail to replicate the stochasticity and complexity of real-world interactions, particularly those involving pedestrians, manually driven vehicles (McCausland, 2019), and nuanced environmental conditions (Chen et al., 2020). The unpredictability of these interactions makes it nearly impossible to simulate every possible corner case. Moreover, relying on detailed cognitive maps for every situation is computationally expensive and often unnecessary (Chen et al., 2015). Instead, we advocate for selectively focusing on safety-critical events, enabling a more efficient and scalable system for risk identification and mitigation.

In addition, while post-accident analyses have traditionally served as the primary source for understanding such failures, they offer limited foresight and often miss an important category of data: near-miss incidents. These non-catastrophic but safety-relevant events—where collisions are narrowly avoided (lytx, 2023)—occur more frequently than actual accidents and thus present a rich, underutilized resource for identifying vulnerabilities in AV behavior. For instance, delayed braking in response to an unexpected pedestrian movement might not result in a crash, but still signals a performance shortfall that merits corrective learning (Walker, 2020). Industry responses such as Cruise's voluntary recall of 300 robotaxis following a low-impact rear-end incident demonstrate the value of proactively leveraging near-miss data to improve safety protocols (Wessling, 2023).

As AV development approaches critical deployment thresholds, Level 2 systems (SAE., 2018)—such as Tesla Autopilot and GM Super Cruise—offer a unique testbed for addressing these limitations. In these systems, human drivers maintain active supervisory control, enabling a form of human-in-the-loop monitoring that not only supports operational safety but also provides critical insight into the AV's perception and decision-making processes. Capturing these cognitive responses—ideally through sensing with minimal user input, which motivates the use of BCIs to unobtrusively record them—offers a novel opportunity for scalable, real-world annotation of potentially hazardous scenarios.

### 4.2 Experimental Protocol for User Onboarding

Calibration onboarding aims to collect individualized templates for labeled cognitive tasks. To prepare users for the upcoming scenario, textual prompts are displayed before the pictorial stimulus, allowing them to initiate the mental imagery process as soon as the scenario appears. The structured calibration process is illustrated in Figure 4.

The protocol begins with a fixation cross at the center of the screen to help users concentrate. Next, a brief textual cue is shown for 2 seconds to inform the user of the upcoming scenario. Immediately after, an image depicting the scenario appears, at which point the user is instructed to mentally simulate the actions they expect the AV to take in response (customized for our studied task). This mental simulation lasts for 5 seconds. Each scenario is presented five times.

This experiment investigates two scenarios as a starting point. To prevent familiarity effects, trials from the two scenarios are randomized in presentation.

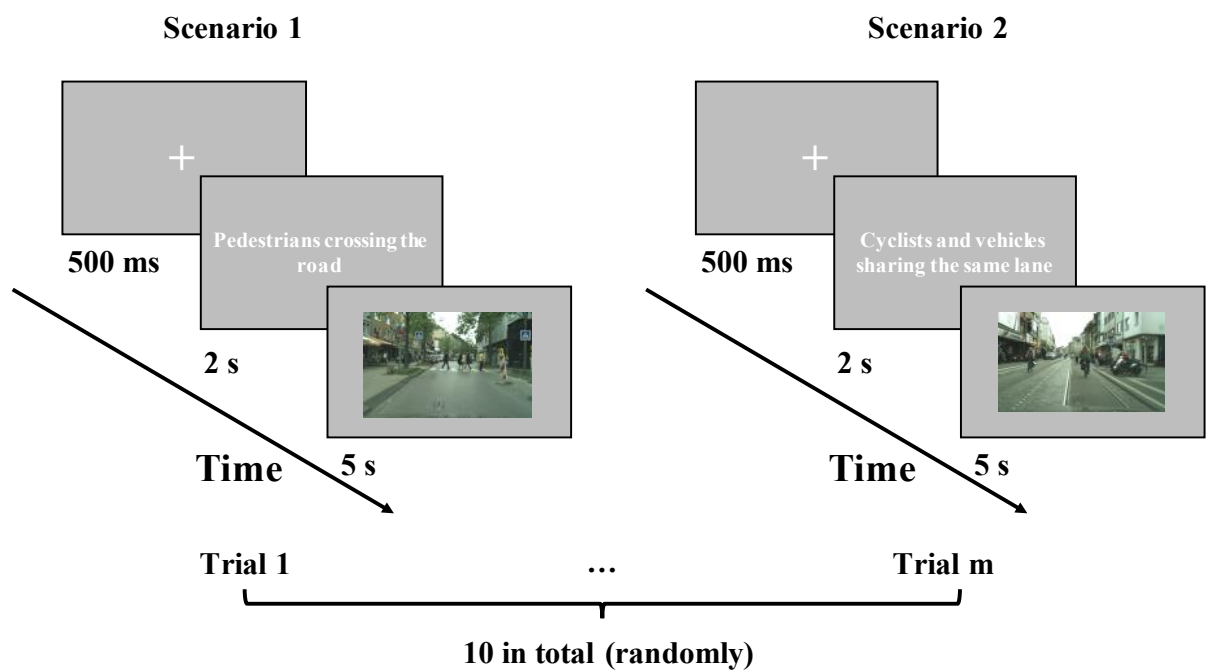

Figure 4. Calibration session protocol.

To test if the protocol works, we also collected five trials of data from the two scenarios while the pictorial stimulus was directly presented without the preceding textual cues. To prevent familiarity biases, the pictorial stimuli used here were different from those used during calibration. Details of the stimuli used in this experiment are provided in Section 4.5.

### 4.3 Platform

The stimuli were presented to the users via the PsychoPy software (Peirce et al., 2019). For event-related data collection, we need to collect event triggers (including labels and timestamps) in addition to the primary data stream of interest. Psychopy has good compatibility with the Emotiv system, which we used to record EEG data. Event markers were synchronized with the recordings (as shown as the vertical dashed lines in Figure 12(a) at the onset of each stimulus. Details on configuring the stimulus presentation order, duration, loop settings, and marker triggers are provided in Figure 5.

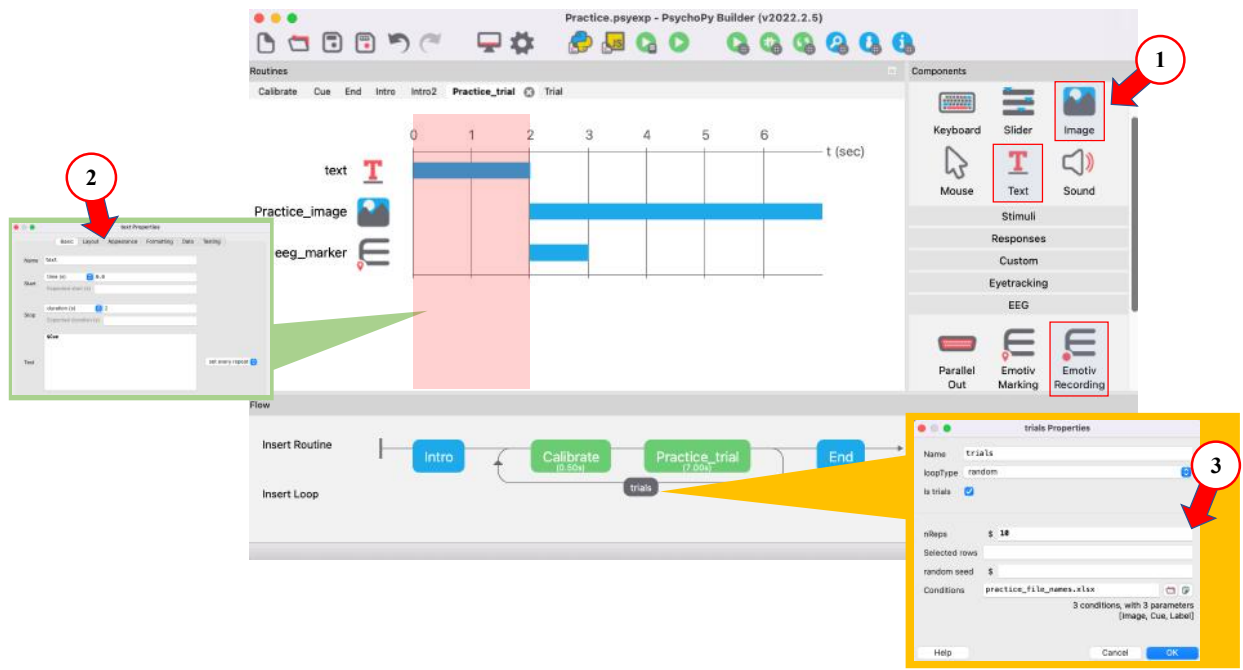

Figure 5. PsychoPy interface for stimulus presentation and synchronization with Emotiv for data recording and trigger marking.

We recorded participants' EEG signals with a 14-channel Emotiv EPOC+ headset, as shown in Figure 6. A 43 Hz low-pass filter was applied to remove high-frequency noise (e.g., electrical line interference). Signals were transmitted to the recording software EmotivPro via Bluetooth.

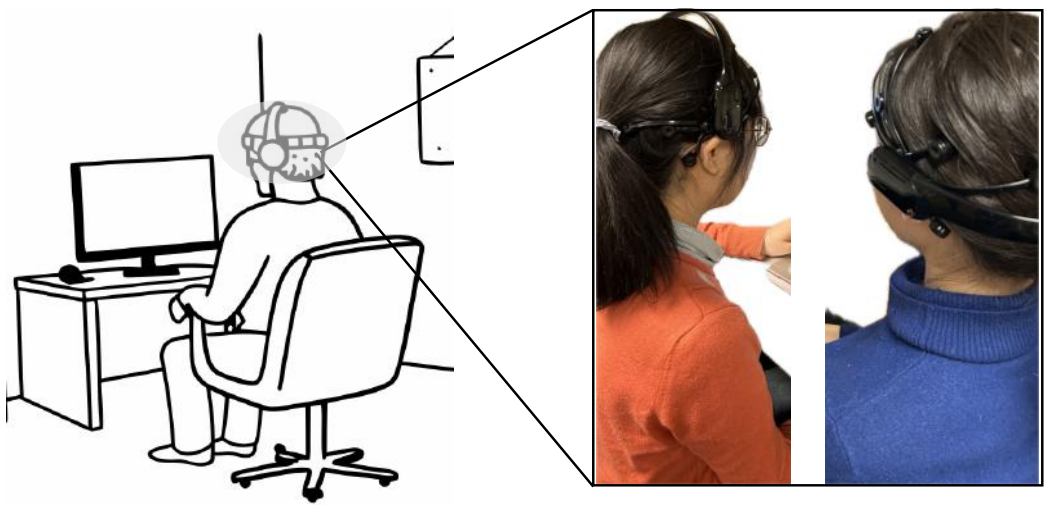

Figure 6. EEG data recording using Emotiv Epoch+.

## 4.4 Mental Imagery Paradigm

The most prevalent paradigm in BCI is MI, where users engage in the mental rehearsal of movements without executing them physically. In addition to neurorehabilitation (Braun et al., 2013), MI has extended to fields such as athletic training (Simonsmeier et al., 2021), dentistry, and pediatric skill acquisition (Jofre et al., 2019; Pichiorri et al., 2015). A substantial body of neuroimaging evidence confirms that mentally simulating motor actions elicits activation patterns that closely mirror those seen during actual movement (Behrendt et al., 2021; Sobierajewicz et al., 2017; Sobierajewicz et al., 2016). Moreover, mental imagery also includes cognitive and sensory simulations such as mental singing, arithmetic, and kinesthetic (somatosensory) imagery (Han-Jeong Hwang, 2014). These activities involve the internal recreation of perceptual or motor experiences and engage brain networks associated with actual perception and action. Using mental imagery for onboarding design in BCI systems has a big benefit: it reduces the need for external stimulation (like flashing lights in Steady-State Visual Evoked Potentials (SSVEP) or specific cues in Event-Related Potentials (ERP)-based BCIs (Li et al., 2013)).

## 4.5 Stimuli

This experiment used the pictural stimuli to prompt participants' mental imagery. Images were selected from an open-source dataset, Cityscapes (Cordts et al., 2016). Images from Cityscape were captured by a camera mounted on a vehicle' windshield and depict real-time driving perspectives on urban and suburban roads (as shown in Figure 7).

Vulnerable Road Users, including pedestrians, cyclists, and motorcyclists, were particularly focused on in this study, as they account for approximately 50% of road traffic deaths globally (Radjou & Kumar, 2018) and are a primary cause in high-profile AV accidents (Othman, 2021). Two scenarios were investigated as a starting point: pedestrian crossings and cyclists' lane-changing behaviors. These two scenarios have been reported as the most common and safety-critical hazards involving VRUs that AVs encounter during road operations (Brill et al., 2024; Nuñez Velasco et al., 2021).

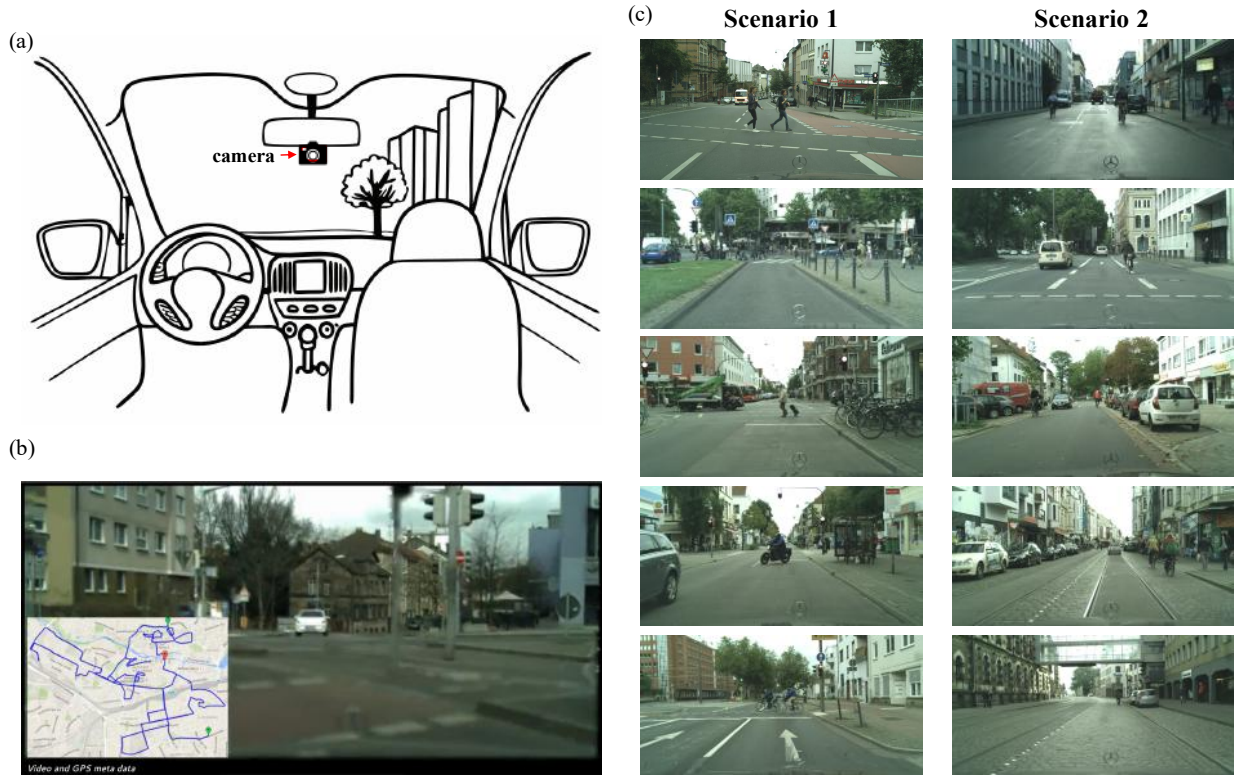

Figure 7. (a), (b) Recording process, and (c) example stimuli used in the experiment.

## 4.6 Cognitive Processing

Cognitive control experiments in psychology typically employ simple visual stimuli (like icons) (as illustrated in Figure 8). This study, however, takes an initial step toward using more complex, less structured scenarios that involve both perception and action. As shown in Figure 9, even seemingly simple visual inputs from real-life can involve complex perceptual processing (Rolls, 2024; Rolls et al., 2022). In our experimental setting, participants were asked to recognize road hazards—pedestrians crossing the road or cyclists sharing the lane (the perception stage)—and then mentally simulate the appropriate responses—braking or steering (the action stage). These mental processes are illustrated in Figure 10. This mental simulation closely mimics the cognitive role of a supervisory driver in Level 2 autonomous driving. Notably, the distinction between perception and action is not strictly separable in mental processing; rather, these stages evolve continuously over time (see Figure 11). However, although the core sequence of cognitive control steps is repeatable, the duration and timing of each stage can vary across trials.

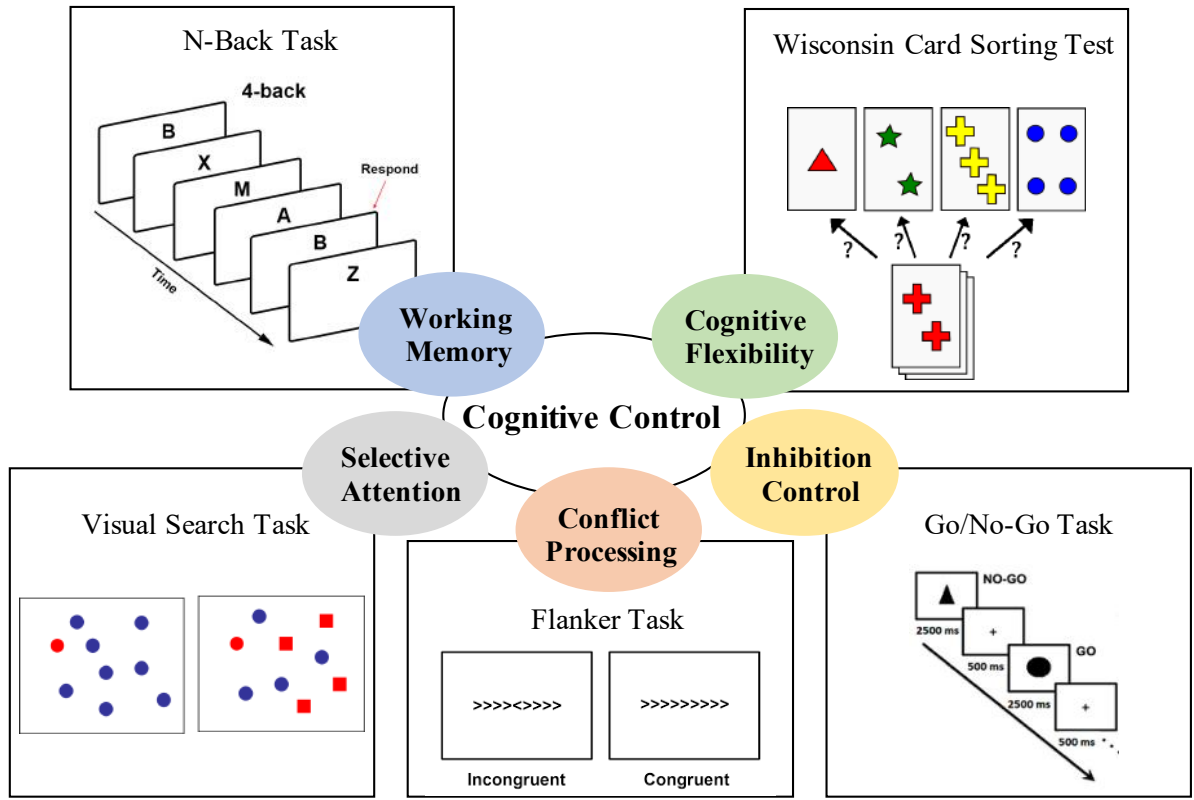

Figure 8. Protocol and stimuli used in psychological tasks to evaluate cognitive control.

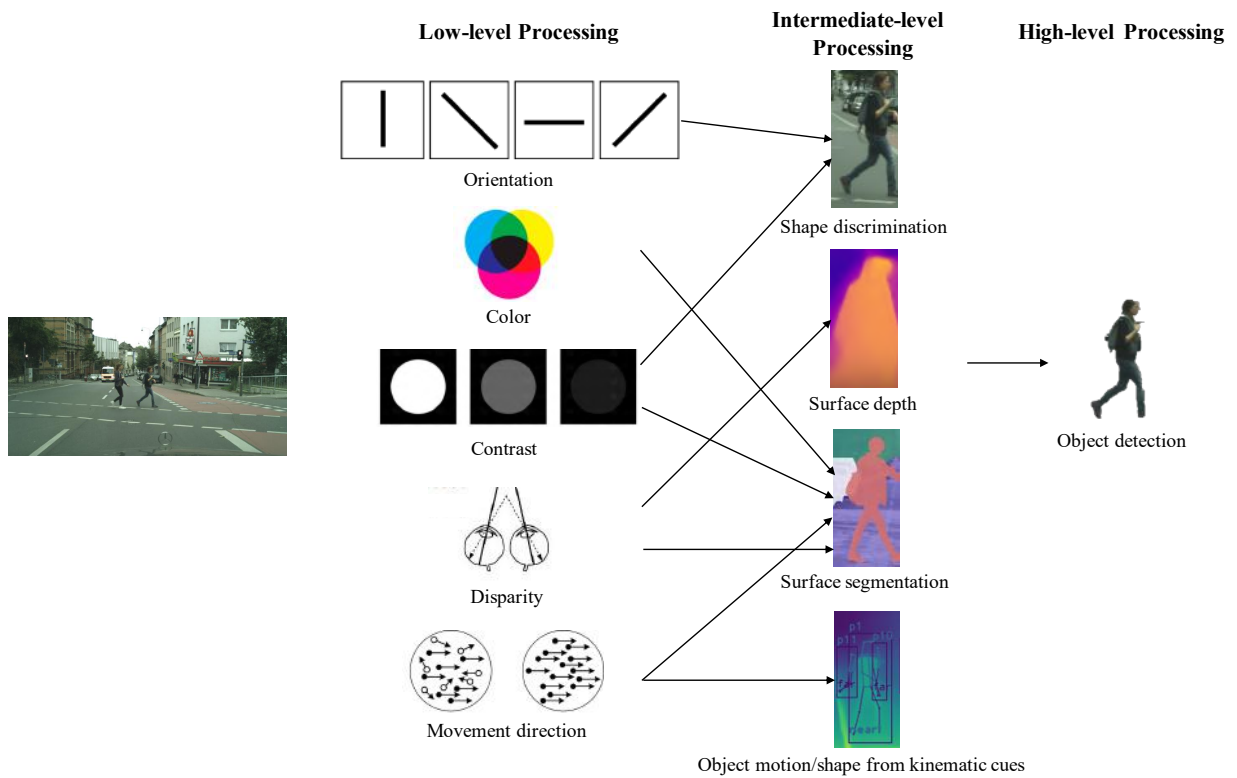

Figure 9. The hierarchy of visual processing for pedestrian recognition from real-world images.

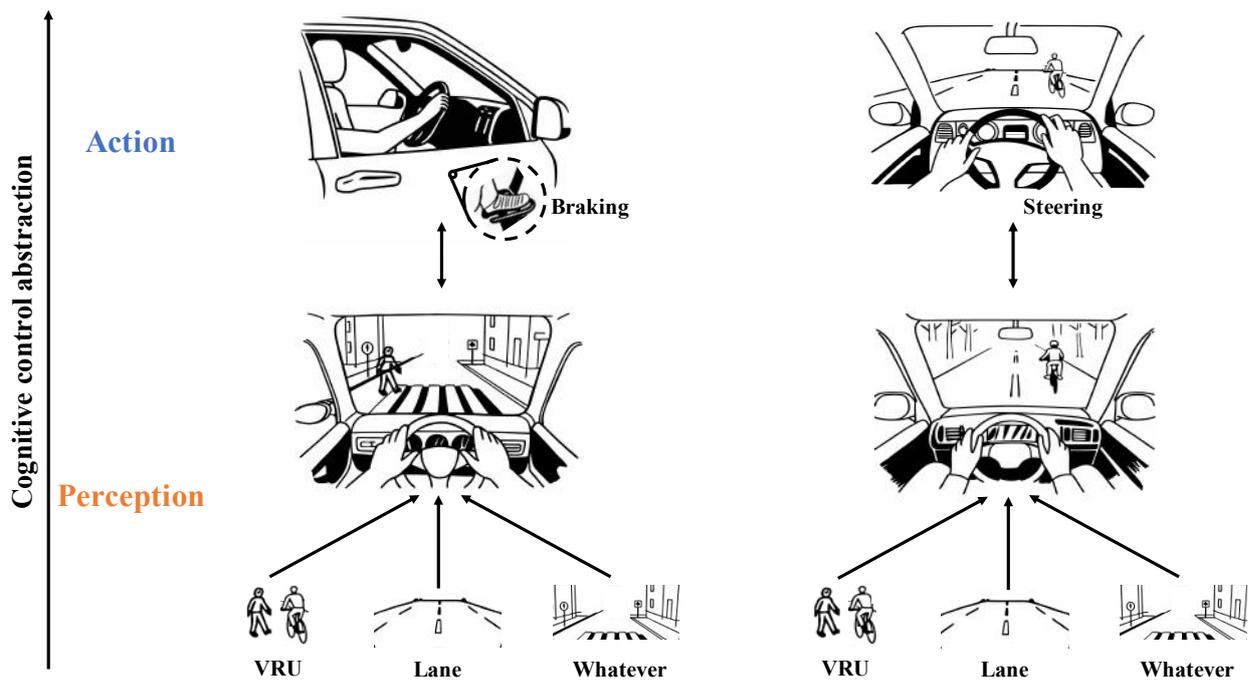

Figure 10. Illustration of the key mental processes for hazard recognition and reaction simulation.

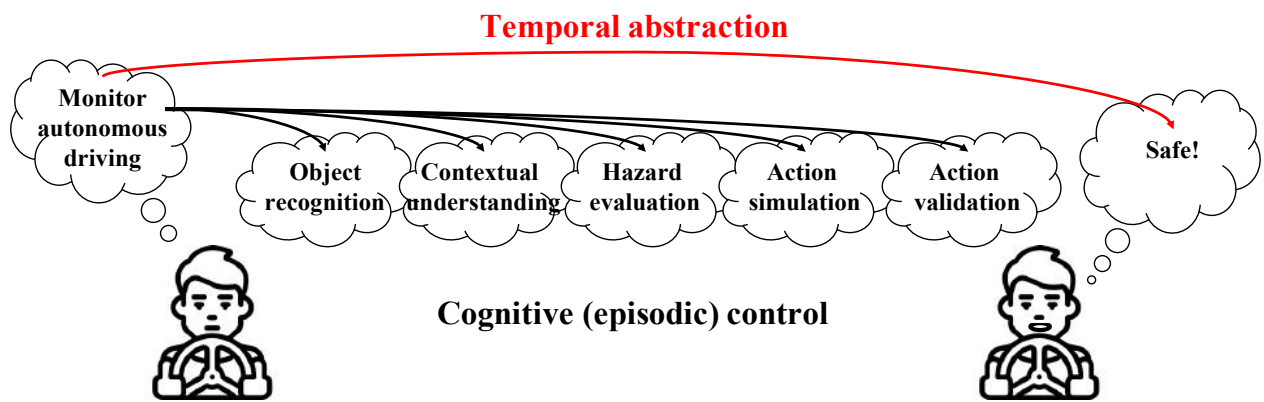

Figure 11. Temporal abstraction of the sequential mental processes in the experiment task.

## 4.7 Participants

Twelve individuals (six males and six females; mean age 24.92 years) were recruited under an Institutional Review Board-approved protocol at the University of Michigan (Reference ID HUM00249262). All participants had normal or corrected-to-normal vision and held a valid driver's license. Based on survey responses during the screening process, all participants reported driving experience, driving at least once per week.

# 5 Results and Analysis

EEG signals were recorded as voltage from 14 channels (arranged as Figure 12(b), which corresponds to the modified International 10-20 system (Klem et al., 1999)). When read as .edf into the MNE-Python package (Gramfort et al., 2013), it appears as Figure 12(a). Power Spectral Density (PSD) was calculated as input EEG features based on Welch's method (Welch, 1967), where signals were first split into overlapping segments, with each segment windowed and processed using the Fast Fourier Transform (FFT). The magnitude squared (power) was then averaged to compute the PSD:

$$PSD(f) = \frac{1}{K}\sum_{k=1}^{K}\frac{1}{NW}\left|\sum_{n=0}^{N-1}x_k(n)w(n)e^{-j2\pi fn/N}\right| \tag{19}$$

where $x_k(n)$ represents the $k$-th segment of the EEG signal (length $N$), $w(n)$ is the window function (Hann window), $W$ is the window normalization factor, $K$ is the number of segments, and $f$ is the frequency bin.

Baseline correction was applied using the time window [-2s, 0s] (0s denotes the stimulus onset). PSD was computed from continuous raw EEG signals (without event segmentation), as shown in Figure 12(c). After segmenting EEG signals into time-locked trials (referred to as epochs), PSD was computed again and aggregated across multiple epochs. The event-related spectral changes are shown in Figure 12(d). Then, EEG power was extracted from five frequency ranges: theta (4–8 Hz), alpha (8–12 Hz), low beta (12–16 Hz), high beta (16–25 Hz), and gamma (25–45 Hz), using a fixed-length Hanning window with a sampling rate of 8 Hz.

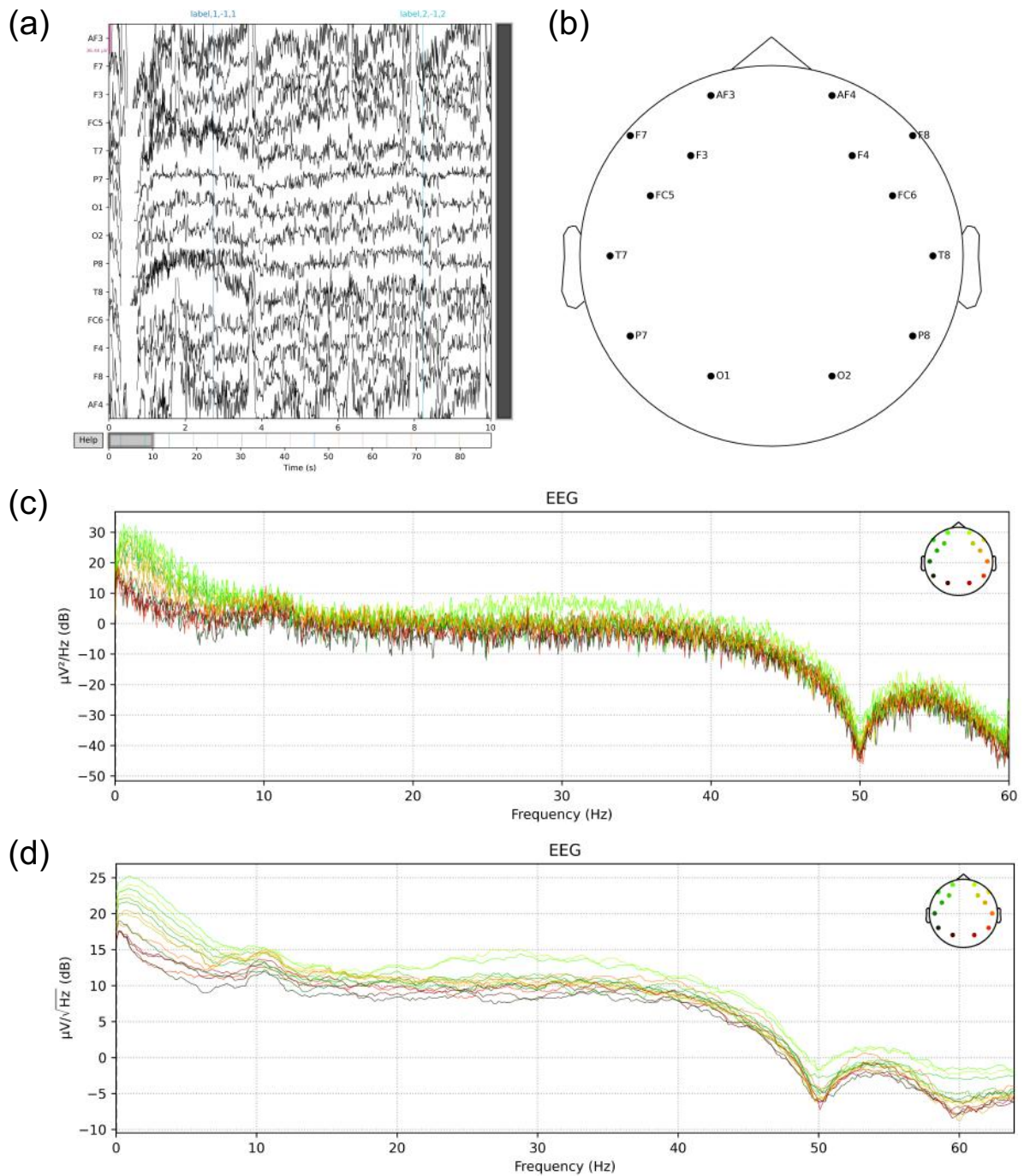

Figure 12. (a) Raw EEG signals; (b) Placement of the 14 sensor channels; (c) Power Spectral Density (PSD) decomposition of continuous signals into frequency components; (d) PSD from evoked responses.

Since the PSD does not contain temporal information, multitaper spectrum estimation was performed using seven Discrete Prolate Spheroidal Sequence windows (Slepian, 1978) to generate time-frequency representation (TFR). The evoked TFRs from two scenarios are shown in Figure 13.

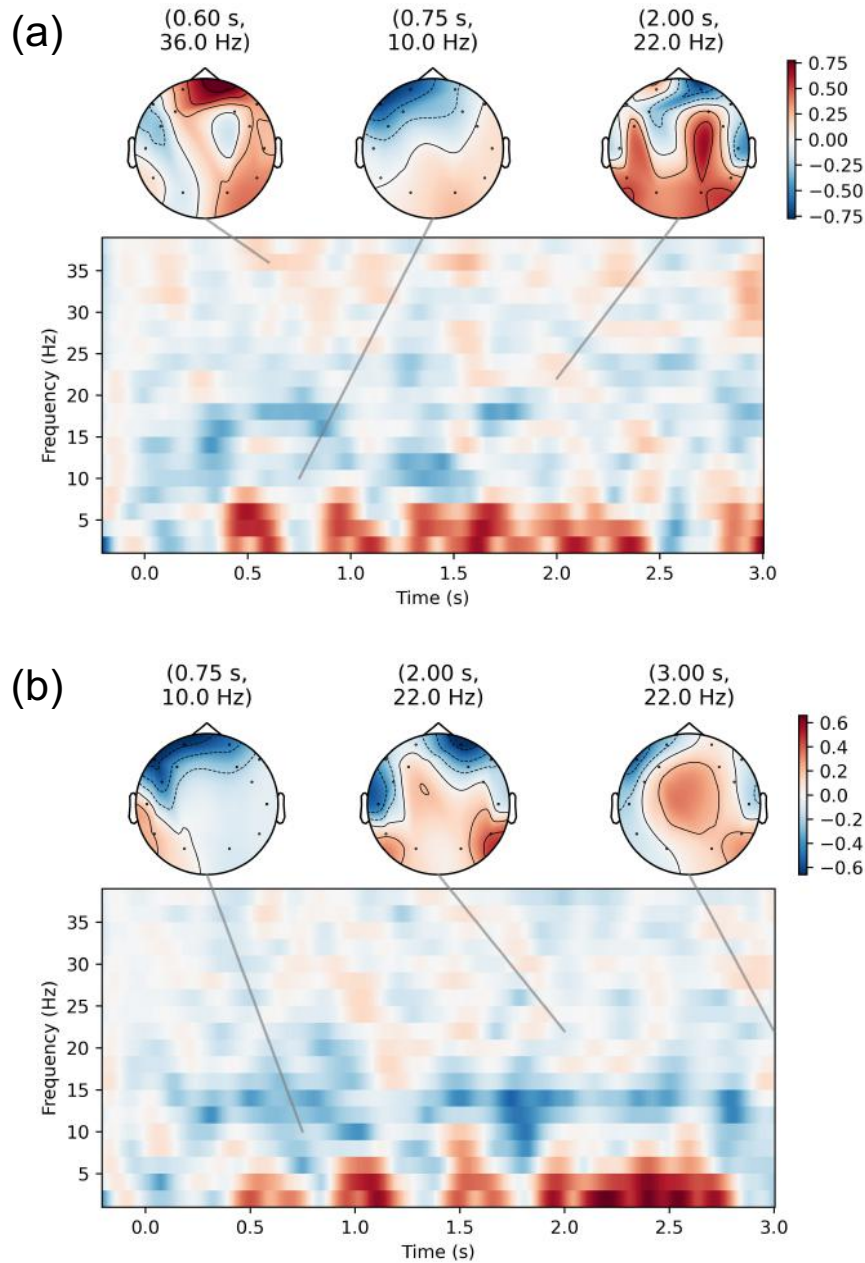

Figure 13. Time–frequency representations (TFRs) from two scenarios: (a) pedestrian crossing and (b) cyclist sharing the same lane.

In Figure 13(a), for the pedestrian crossing and braking mental imagery task, gamma activation is predominantly observed in the centro-right frontal region. This finding aligns with previous neurophysiological studies that report increased right frontal gamma activity in preparation for a stopping action (Swann et al., 2012). Research also suggests that during voluntary movement, gamma-band activity in the anterior supplementary motor area (SMA) increases. This activity can be detected non-invasively using EEG over the frontal midline, likely playing a role in mediating upcoming actions (Ball et al., 2008). Additionally, alpha-band ERD was observed over the left frontal area (at 0.75s). This neural signature has been reported in previous studies as indicative of enhanced cognitive control and activation during cognitive reappraisal of visual stimuli (Parvaz et al., 2012). It is also likely to be modulated during motor preparation (Platz et al., 2000). Moreover, evidence from the third topomap indicates that MI is predominantly characterized by beta-band activity in the sensorimotor cortex. This finding is consistent with prior findings showing an increase in beta-band power in the pre-SMA at the time of stopping (Swann et al., 2012).

Comparing the oscillatory effects between the two scenarios (Figure 13(a) vs. Figure 13(b)), a theta phase delay was observed in the lane-changing mental imagery task. Previous studies suggest that theta activity is linked to behavioral inhibition (the ability to wait before acting) (Fakhraei et al., 2021). Theta oscillations have also been shown to emerge around behavioral decision points, where they play a key role in evaluating choice-relevant information. This is supported by connectivity evidence indicating an association between theta activity and visuospatial memory in the frontoparietal cortex (Rawle et al., 2012). Considered together, these findings suggest that the braking action can be more intuitive and reflexive, while the lane-changing action likely involves greater reasoning and recall. Furthermore, connectivity analyses indicate that theta frontocentral connectivity gradually decreases during movement planning, reflecting the progressive disengagement of frontal executive control after its initial role in suppressing established visuomotor maps (Gentili et al., 2015).

To further validate this, we also analyzed functional connectivity using weighted Phase Lag Index (wPLI) (Vinck et al., 2011):

$$wPLI = \frac{\sum_t |Imag(S_{xy}(t))| \cdot sign(Imag(S_{xy}(t)))}{\sum_t |Imag(S_{xy}(t))|} \qquad (20)$$

where $S_{xy}(t)$ is the cross-spectral density (Morlet wavelets of signals $x(t)$ and $y(t)$). $Imag$ extracts the imaginary component of the cross-spectrum, and then $sign$ determines the sign of the imaginary part.

Figures 14(a) and (b) show sensor connectivity at global theta wPLI peaks, with a maximum at 0.562s and a minimum at 1.219s post-stimulus. We observed that frontocentral connectivity weakened over time, supporting the idea that theta activity modulates motor planning and can serve as a marker of motor preparedness.

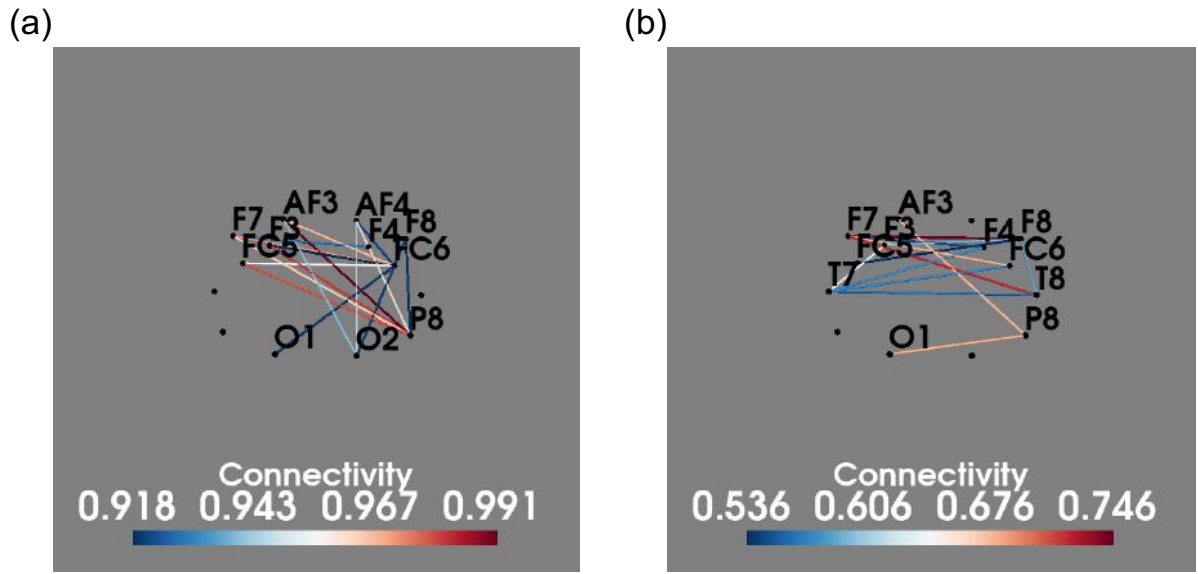

Figure 14. Theta-band sensor connectivity at (a) maximum and (b) minimum values of the weighted Phase Lag Index (wPLI).

Since EEG is recorded from the scalp, the measured signals are spatially smeared representations of underlying neural activity. To better identify the true cortical sources that generate the observed scalp signals, we performed source localization using the MNE-Python toolbox. The steps were briefly described as follows. Algorithms and mathematical details can be found in MNE (2025).

1) Forward Model Construction

A three-layer boundary element model (BEM) was created based on the MNI-152 standard brain, using FreeSurfer-generated anatomical surfaces. The head model defines the physical geometry (the conductivity boundaries) for projecting EEG data onto cortical surfaces.

2) Sensor Coregistration

EEG sensor locations were aligned with the template head surface using fiducial markers (nasion, left/right preauricular points) and digitized sensor positions to ensure anatomical accuracy.

3) Source Space Setup

A cortical source space was defined with a spacing of approximately 5 mm between dipoles. This discretizes the brain volume into thousands of candidate source locations.

4) Noise Covariance Estimation

Noise covariance was estimated from a baseline ([-0.2s, 0s] pre-stimulus) time window to regularize the inverse solution and account for background activity.

5) Inverse Solution Computation

We applied the Minimum Norm Estimation (MNE) inverse method to estimate distributed source activation from the sensor-space evoked responses.

The source localization results are shown in Figure 15. Two dominant source activations were identified during user calibration. In the left hemisphere, the source MNI coordinates were found to be [-53.28, -20.34, 21.85]. This source localizes near the left inferior frontal gyrus (pars opercularis), corresponding to Brodmann area 44. This region is classically part of Broca's area, which is established to correlate with language processing (Binder et al., 1997), particularly syntactic structure, inner speech, and subvocal rehearsal. The involvement of this region suggests that participants may be engaging in verbal or symbolic internal representations (e.g., self-instruction, mental labeling) during mental imagery. In the right hemisphere, the source MNI Coordinates were [27.79, -0.80, 46.93]. This activation localizes near the right middle frontal gyrus, associated with Brodmann area 9. This area is a key part of the dorsolateral prefrontal cortex, implicated in executive control, working memory, and goal maintenance (Curtis et al., 2000). Its engagement aligns well with the cognitive control demands of our BCI tasks, particularly when users must maintain and update mental states (e.g., mental imagery vs. resting) and perform top-down modulation of sensorimotor rhythms (SMRs).

Overall, these findings indicate that the left inferior frontal gyrus activation may reflect the use of language-like cognitive scaffolds during mental rehearsal, while the right dorsolateral prefrontal cortex supports the maintenance and monitoring of task goals. These regions may also contribute to intra-individual consistency in EEG patterns, enabling more robust few-shot classification.

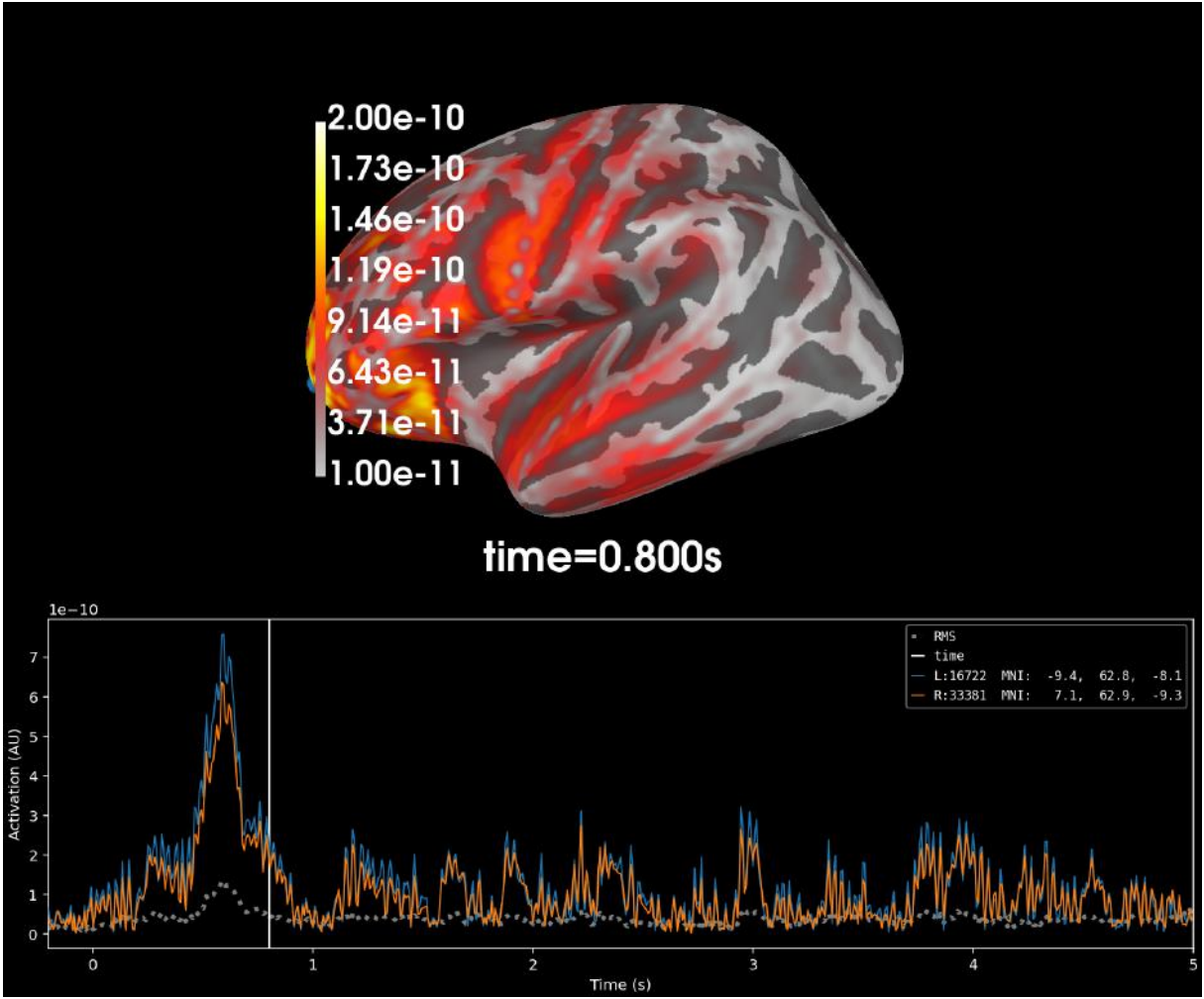

Figure 15. Source localization results showing estimated cortical activation based on evoked EEG data during calibration.

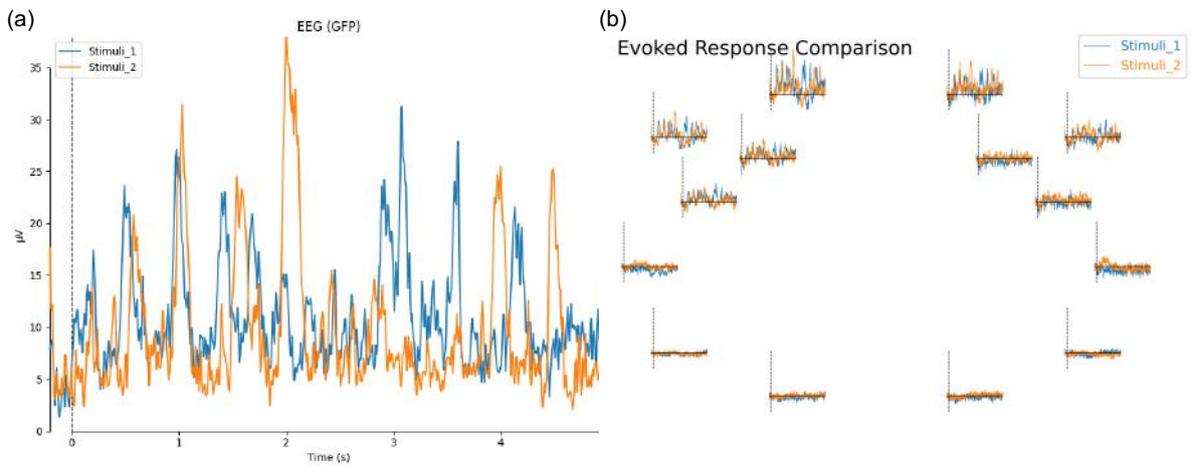

Figure 16. Evoked response comparison shown in (a) global field power and (b) across all channels.

Figure 16(a) shows the global field power (Lehmann & Skrandies, 1980, 1984; Murray et al., 2008), representing signals picked up by all sensors across the entire scalp under two scenarios:

$$GFP(t) = \sqrt{\frac{1}{N}\sum_{i=1}^{N}(V_i(t) - \bar{V}(t))^2} \tag{21}$$

where N is the number of channels, $V_i(t)$ is the voltage at channel $i$ at time $t$, and $\bar{V}(t)$ is the average voltage across all channels at time $t$. Mathematically, the GFP is the population standard deviation across all sensors, calculated separately for every time point. If all sensors have the same value at a given time point, the GFP will be zero at that time point. If the signals differ, the GFP will be non-zero at that time point.

Figure 16(b) displays the evoked response comparison in a topographical layout, with each small subplot corresponding to an EEG channel. Some channels show more noticeable differences in amplitude or shape (e.g., peak latency) between the two scenarios, particularly those located in the frontal and parietal regions. In contrast, the posterior occipital region shows less divergence. According to the visual pathways, the posterior brain regions are primarily responsible for low-level visual processing (such as edge detection, motion, and contrast; see Figure 9), and may therefore share similar activation patterns across the two scenarios.

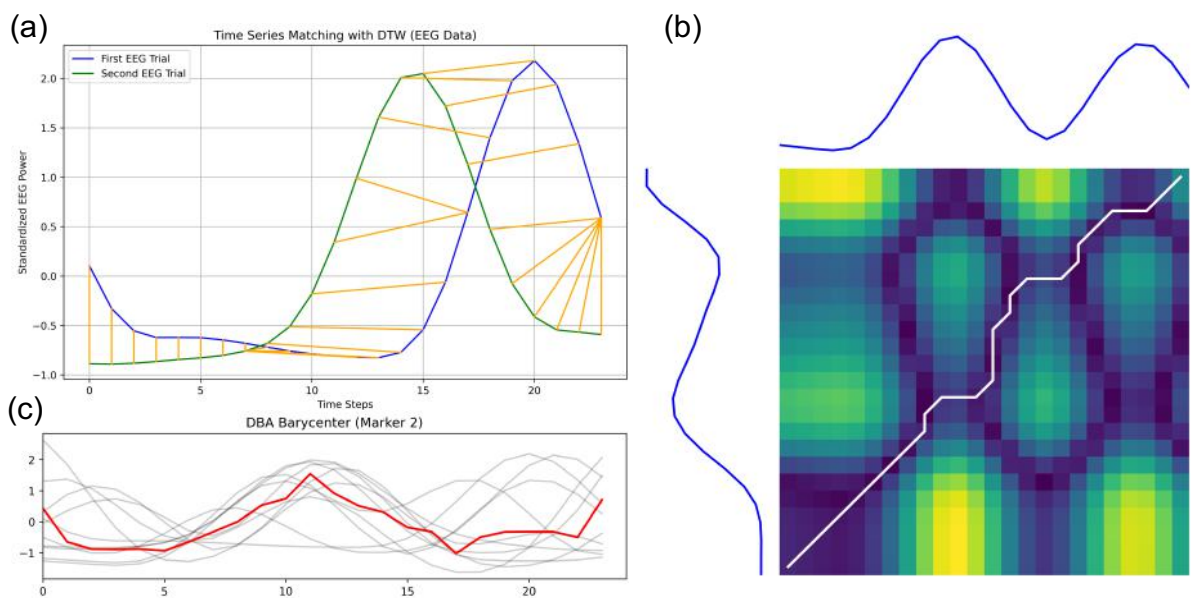

Figure 17. (a) Illustration of time point alignment between two EEG sequences using the Dynamic Time Warping (DTW) method; (b) Optimal warping path identified through the cost matrix; (c) Illustration of the Dynamic Time Warping Barycenter Averaging (DBA) computed from multiple trials.

To identify the channels and features that best distinguish between the two scenarios (pedestrians crossing the road or cyclists sharing the lane), DTW distances were computed. The results highlighted AF4.Theta, AF3.Theta, F8.Theta, and F7.Theta (in descending order) as the most discriminative features. Figure 17(a) illustrates the DTW alignment process between two EEG trials. The optimal warping path through the cost matrix is shown in Figure 17(b). Subsequently, the most discriminative feature was used to compute the DBA barycenter as a template for similarity comparison, as demonstrated in Figure 17(c). Although multiple trials go through repeatable mental processes, the timing of those processes can vary (i.e., phase delays). The DBA technique is thus designed to account for these temporal shifts when averaging time series.

Our Siamese Dual-Attention Convolutional Network achieved up to 80% accuracy on individualized datasets for classifying two mental imagery tasks in response to two road hazards. This performance was obtained under a five-shot learning setting and exceeds the widely accepted 70% threshold for binary classification viability in BCI applications that indicates practical feasibility (Power et al., 2012). To further assess the effectiveness of the feature selection process and the performance of the classifier, we conducted two additional evaluation tests.

First, we used the same Siamese network as the feature extractor; however, instead of feeding in the EEG power dimensions identified during the feature selection step, we randomly selected features from the remaining dimensions. This resulted in a substantial drop in classification accuracy, with an average accuracy of only 20%. The outcome suggests that the network was effectively trained to represent the selected features, and that random features do not carry meaningful discriminative information.

Second, we used Integrated Gradients (IG) (Sundararajan et al., 2017) as an explainable technique to attribute the model's predictions to its input features. The formula for calculating IG is as follows:

$$IG_i(x) = (x_i - x_i') \cdot \int_{\alpha=0}^{1} \frac{\partial F(x' + \alpha(x - x'))}{\partial x_i} d\alpha \qquad (22)$$

where $i$ is the feature index, $x$ is the input, $x'$ is the baseline, and $\alpha$ is the interpolation constant used to perturb the input. Since computing the definite integral directly is often numerically impossible and computationally costly, we instead approximate the integral using a Riemann sum (Hughes-Hallett et al., 2020). This gives:

$$\int_0^1 \frac{\partial F(x_\alpha)}{\partial x_i} d\alpha \approx \frac{1}{m} \sum_{k=1}^{m} \nabla F(x_\alpha) \qquad (23)$$

This analysis was conducted using the TensorFlow package in Python (TensorFlow, 2024). To evaluate feature importance, we performed a feature ablation test. Figure 18 presents the IG results for models with and without the dual-attention modules, using a 3-second time window and a sampling rate of 8 Hz. Overall, the model with the attention module demonstrates greater sensitivity to temporal dynamics, with important features exhibiting stronger influence at specific, meaningful time steps. Both models successfully capture frontal theta activity—associated with motor readiness—and synchronized sensorimotor high beta activity—linked to MI. However, in the attention-based model, these discriminative features exert a more pronounced effect on the model's predictions. Simultaneously, features that likely play a shared or non-discriminative modulatory role across both scenarios are more effectively suppressed in the attention-based model. This suggests that the attention mechanism enhances the model's ability to focus on context-relevant features while minimizing the contribution of less informative ones.

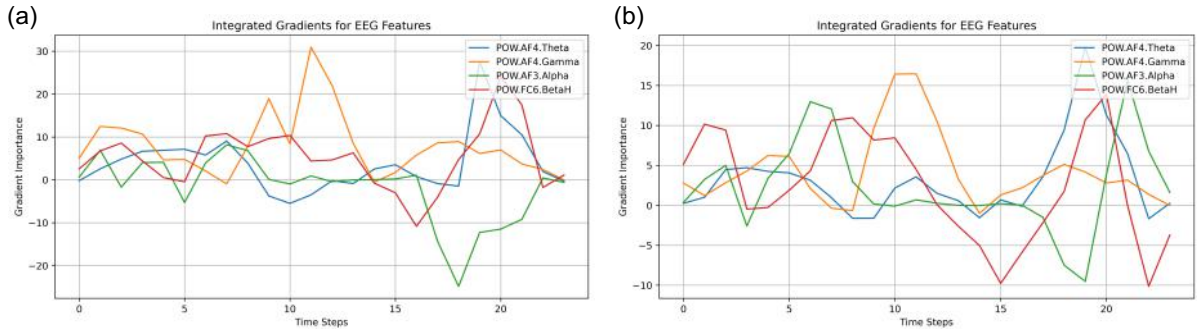

Figure 18. Integrated Gradients calculated from representative feature dimensions using the Siamese Network: (a) with dual attention and (b) without attention.

## 5 Theoretical Discussion

### 5.1 Learning and Memory

The core mechanism of forming templates for similarity comparisons in our BCI system resonates with the Complementary Learning Systems perspective (McClelland et al., 2020), which posits that experiences are stored in distributed and complementary memory systems. In machine learning, this memory-like capability is also being explored via differentiable memory-augmented controllers to enable neural networks to remember and retrieve information flexibly. For example, (Pritzel et al., 2017) introduced a deep reinforcement learning agent equipped with a Differentiable Neural Dictionary (DND), which stores state–action–value tuples and supports decision-making by retrieving relevant past experiences via associative memory mechanisms. Similarly, (Lu et al., 2022) incorporated a memory module into a neural network to enhance the prediction of future environmental states, while also introducing selective encoding and retrieval to prioritize the most relevant stored information.

Inspired by these approaches, our system introduces the BCI agent that perceives the environment through a Siamese network acting as a feature extractor. Given EEG observations $x_t$, the network computes an embedding vector $h$ using DBA. This embedding serves as a key for memory storage within a DND-like structure. Specifically, $h$ is added to a key matrix that stores representative templates for each event or mental category, enabling flexible retrieval and comparison during future interactions. This storage can be described as:

$$H_k = DBA(\{h_i | y_i = k\}) \tag{24}$$

where $H_k$ is the representative template embedding for event category $k$. The reference embeddings can be adapted by recomputing the DBA whenever new trials are added:

$$H_k^{(new)} = DBA(\{h_i | y_i = k\} \cup h_{new}) \tag{25}$$

The biological correspondence of this line of machine learning research—including our own—can be found in the prefrontal–hippocampal circuit, which plays a central role in memory encoding and retrieval (Pilly et al., 2018).

In contrast to the often rigid and deterministic retrieval mechanism defined in machine learning models, biological memory retrieval is inherently probabilistic and dynamic. We thus propose to incorporate more biologically plausible retrieval dynamics. In particular, we suggest the BCI agent advance toward architectures that more closely mirror human cognition, such as the ACT-R cognitive architecture (Ritter et al., 2019). In ACT-R, memory retrieval is influenced by mechanisms analogous to the Ebbinghaus Forgetting Curve (Ebbinghaus, 2013), where the accessibility of a memory trace decays exponentially over time. Accordingly, when relying on stored templates to classify EEG signals, the older templates are weighted less:

$$w_i = \frac{k(H_i, q)}{\sum_j k(H_j, q)} \cdot e^{-\lambda t_i} \tag{26}$$

However, when users are actively engaged with the BCI system, the strength of memory templates will be reinforced upon each retrieval. This idea reflects principles from Hebbian learning ("neurons that fire together wire together") (Pulvermüller, 1996, 2001) and spaced repetition (Smolen et al., 2016), where repeated use strengthens memory. Such a process can be described as:

$$P_{retrieval}(t, r) = e^{-\lambda t} + \beta r \tag{27}$$

where $t$ is the time elapsed since storage, and $r$ is the retrieval frequency. Therefore, in our system, although templates naturally decay over time due to storage constraints, retrieving them can refresh and strengthen them. Frequently used templates become more robust. This kind of dynamic memory management could be key to building a generalized BCI system, especially when the user performs multiple tasks, each with multiple categories that must be classified. Further discussion on multi-class classification appears in Section 5.3.

The template matching process (searching and selecting the right template) is the core operation of the system and demands a form of executive control—in humans, this is the ability to follow and regulate mental procedure. This ties closely to working memory and the global workspace theory, which sees the brain as having a temporary storage buffer (like a scratchpad) that coordinates the flow of conscious information across specialized brain modules. The global workspace is like a "router" or "signalman": it determines what information gets processed, in what order, and by which systems. But the operation within this workspace serial and slow—one thing at a time—which psychologists call the "central bottleneck" (Tombu & Jolicoeur, 2005). Thus, if the BCI seems unresponsive, it might be encountering a similar bottleneck. But unlike the human brain, machines could potentially overcome this limitation with greater parallelism and computational power, which is a unique potential for artificial agents to exceed the limits of human cognitive throughput.

### 5.2 Adaption and Attention

This subsection explores how transfer learning, continual learning, and federated learning—in conjunction with attention mechanisms—can help enhance the adaptability of BCI systems across individuals and tasks.

A central challenge in EEG-based BCIs is the neurophysiological variability underlying SMRs, which fluctuate both over time (intra-subject variability) and across users (inter-subject variability). These fluctuations result in covariate shift, a change in the statistical properties of the input feature (EEG data) distribution, so a model trained on one person/session may not work well on another. Transfer learning addresses this problem by adapting pre-trained models to new domains (Saha & Baumert, 2020). For instance, (Fahimi et al., 2019) applied transfer learning to adapt CNN-learned EEG feature representations across subjects for improved inter-subject generalization in BCI applications. In our framework, we apply a similar concept by using a Siamese network pretrained on data from one individual and fine-tuning it on a new user. The fine-tuning process for individual $j$ is expressed as:

$$\theta_j = \theta^* - \eta \nabla_\theta \mathbb{E}_{(z,x,y)}[L(y, f(z, x; \theta))] \qquad (28)$$

To prevent catastrophic forgetting when models are sequentially exposed to new tasks, continual learning has previously been investigated (Zhou & Liao, 2023b). A regularization term is introduced to preserve prior knowledge:

$$\theta_t = \theta_{t-1} - \eta \nabla_\theta (\mathbb{E}_{(z,x,y)}[L(y, f(z, x; \theta))] + \lambda \Omega(\theta) \qquad (29)$$

where $\Omega(\theta)$ is a knowledge retention penalty that prevents drastic weight changes. Different continual learning techniques define $\Omega(\theta)$ differently. For example, in Elastic Weight Consolidation (Kirkpatrick et al., 2016):

$$\Omega(\theta) = \frac{1}{2} \sum_i F_i (\theta_i - \theta_i^*)^2 \qquad (30)$$

where $F_i$ is the Fisher information matrix that measures the importance of each parameter.

Some works use the replay-based methods, where past EEG data and stored and mixed with new data:

$$X_{train} = X_{new} \cup X_{old} \qquad (31)$$

However, replay-based continual learning becomes increasingly memory-intensive as more users join. This limitation motivates the use of federated learning, which allows decentralized learning across many individuals without centralizing the data. In a federated framework, each individual $i$ maintains a local model $\theta_i$, fined-tuned individually. These updates are then aggregated into a global model via federated averaging: $\theta = \sum_{i=1}^{N} \frac{n_i}{N} \theta_i$ \qquad (32)

To address variability across individuals while minimizing retraining costs, we introduce a selective adaptation strategy: base convolutional layers are frozen, while attention modules remain trainable. The Squeeze-and-Excitation module enables channel-wise attention, learning task-specific feature importance. Meanwhile, the Multi-Head Self-Attention mechanism dynamically adjusts temporal focus $softmax(QK^T)$. Thus, this strategy uses localized updating of attention modules while retaining general feature extractors and output heads. The use of attention mechanisms is particularly promising when extending to multi-task BCI scenarios, where attention modules can support task-specific adaption without overhauling the entire model architecture:

$$\theta_{attn}^{(t+1)} = \theta_{attn}^{(t)} - \eta \nabla_{\theta_{attn}} \mathbb{E}_{(z,x,y)}[L(y, f(z, x; \theta_{conv}^{frozen}, \theta_{attn}, \theta_{dense}^{frozen}))] \qquad (33)$$

This approach aligns with recent developments in federated transfer learning, where instead of aggregating multi-subject datasets centrally, models extract shared discriminative representations using domain alignment technique (Ju et al., 2020). In line with this, (Cai et al., 2023) proposed a hierarchical domain adaptation framework that maps EEG signals from various sources into aligned subspaces using projective dictionary learning, while preserving local discriminative structures.

Interestingly, the foundational concept of federated learning resonates with insights from cognitive neuroscience. Human learning is accelerated by social attention sharing (Dehaene, 2020), a phenomenon confirmed by developmental psychology. For example, infants learn new words or phonemes effectively only when engaged in shared attention (Kuhl et al., 2003). In BCI, despite variability in EEG signals across individuals, many BCI studies have shown consistent neural patterns for the same tasks, providing a foundation for cross-subject weight transfer in federated learning. This form of biologically inspired distributed learning may significantly accelerate deployment and personalization of BCI agents while preserving individual data privacy.

## 5.3 From Binary to Multi-Class Classification

Retrieval from the stored EEG reference set is proposed to follow a process analogous to nearest-neighbor search in DND. While binary classification relies on direct comparisons between two categories, we can extend this framework to the multi-class setting by introducing a distance-based kernel function $g$ to compute similarity between the input EEG embedding and stored reference templates. Given a query EEG epoch $q$:

$$g(H_k, q) = \frac{1}{\|K_i - q\|^2 + \delta} \tag{34}$$

where $\delta$ is a small stabilizing term.

To improve classification robustness, we draw inspiration from associative memory mechanisms in Hopfield networks (Millidge et al., 2022). Specifically, we implement a separation function that ensures only the most relevant EEG templates influence the decision. This function identifies the top-k most similar stored EEG embeddings, normalizes their similarity weights, and aggregates their contributions to produce a soft classification decision. Mathematically, it is defined as:

$$sep(x, k) = \frac{\kappa(x, k)}{\sum_i \kappa_i(x, k)} \tag{35}$$

where:

$$\kappa_i(x, k) = \begin{cases} x_i, & if\ x_i\ is\ among\ the\ top\ k\ matches \\ 0, & otherwise \end{cases} \tag{36}$$

Then, the retrieved memory representation is:

$$z = \sum_i h_i \kappa_i \tag{37}$$

# 6 Conclusion

This study presents a cognitively inspired BCI agent capable of classifying distinct mental activities with minimal training examples. By leveraging similarity learning and a Siamese neural network framework, the agent learns to form adaptable mental schemas, analogous to the way humans—especially infants—learn to categorize and generalize across variable sensory inputs. Our model currently achieves 80% classification accuracy in a few-shot setting, and interpretability analysis using IG confirms that the neural network captures meaningful features for prediction. These results demonstrate strong early-stage potential for distinguishing between different mental imagery categories. In the experimental tasks, we are actively expanding the model's capacity to handle more complex and dynamic imagery that reflects real-world cognitive processes. This marks a significant advancement toward developing BCIs that are both scalable and adaptable to naturalistic environments. In the broader context of cognitive development and human-machine interaction, we view this work as an early but significant step toward equipping artificial agents with mechanisms for meta-learning, i.e., learning to learn. From the developmental perspective, just as the human prefrontal cortex matures over years through experience and feedback, we expect our BCI agent to evolve similarly with the proposed internal cognitive architecture that is enhanced by mechanisms for attention and memory.

This pathway for collecting human feedback is important because intelligence in robotics, such as AVs, requires not only technical capabilities like path planning and obstacle avoidance, but also high-level situational awareness. Some hazards—like so-called "invisible hazards" or unexpected dangers—are not easily detectable by sensors but can lead to catastrophic consequences or near misses (lytx, 2023). These rare situations are difficult to capture in training data or replicate in simulations, because they are statistically rare (the so-called long-tail problem), and thus hard to model using current data-driven or probabilistic large language model-based approaches (Zhang et al., 2024). However, once AVs are deployed in the real world, this research opens the opportunity to collect feedback directly from human drivers, through biological signals such as EEG that is featured by near-simultaneous responsiveness. Building on the proof-of-concept demonstrated in this study, an immediate and practical application of the proposed BCI system is to support human safety supervisors in monitoring AV operations in real time (Chu et al., 2023). In this setting, the BCI passively captures neural responses to decode perceived risks and allow the system to flag safety-critical events that might otherwise go unnoticed by conventional sensors. Over time, this approach can be scaled beyond professional safety drivers to involve everyday drivers, forming the basis for a large-scale, community-driven data collection initiative. In this expanded model, drivers equipped with BCIs contribute cognitive feedback as they interact with AVs, creating a distributed, crowdsourced platform for identifying rare, ambiguous, or safety-relevant scenarios (Kalra &

Paddock, 2016). This not only addresses the long-tail problem in AV training but also establishes a new paradigm for collaborative intelligence between humans and machines.

As robots become more integrated into domestic and assistive settings, humans will naturally seek more intuitive ways to communicate with them, particularly for conveying intentions and preferences. BCIs are uniquely positioned to support this interaction. However, current BCI signal decoding techniques are not well-adapted to complex real-life perception-action tasks, and supervised learning methods require large labeled datasets and long calibration times—barriers to real-world usability. Therefore, we propose better (more simplified and guided) first-time user calibration tools, including intuitive user interfaces (UI), and call for deeper investigation into cognitive architectures that enable social learning from human counterparts in robotic agents. For future research directions, we encourage researchers to explore knowledge transfer from humans to robots—through expert systems, reinforcement learning with human feedback, imitation learning, and transfer learning frameworks.

In addition, there is a growing need to explore more types of sensory modalities for evoking mental imagery in BCI research. Results from our EEG source localization show activation in the language-related Broca's area, suggesting that reading sentences or watching actions (i.e., linguistic or motion-based stimuli) might also work well—or even better—for engaging cognitive processes related to action simulation. Although a substantial body of neuroimaging literature shows that reading, observing, and imagining actions activate overlapping brain areas (Behrendt et al., 2021; Boulenger et al., 2006; Sobierajewicz et al., 2017; Sobierajewicz et al., 2016), it is beneficial to collect empirical evidence to investigate which stimulus mode (or combination) can help users produce more consistent and robust brain signal patterns. Lastly, the possibility of such "mind reading" certainly raises ethical concerns about privacy and the potential for misuse of such technology. Therefore, it is essential to develop corresponding safeguards, including privacy-preserving techniques and secure data infrastructure, to ensure the responsible use of this emerging technology.

**Conflicts of Interests**

None

**Data Availability Statement**

Some data, models, or code that support the findings of this study are available from the corresponding author upon reasonable request, including the individual EEG data collected during the experiment and the codes for Siamese network.

**Acknowledgements**

The work presented in this paper was supported financially by the United States Department of Transportation Center for Connected and Automated Transportation (CCAT) via Award# 69A3552348305 and by the United States National Science Foundation (NSF) via Award# SCC-IRG 2124857. The support of the CCAT and NSF is gratefully acknowledged. Any opinions and findings in this paper are those of the authors and do not necessarily represent those of the CCAT or the NSF.

Lutz, A. (2002). Toward a neurophenomenology as an account of generative passages: a first empirical case study. *Phenomenology and the Cognitive Sciences, 1*(2), 133-167.

lytx. (2023). Why Near-Collision and Near-Miss Analysis Matters. Retrieved from https://www.lytx.com/blog/why-near-collision-and-near-miss-analysis-matters#:~:text=What%20is%20a%20near%20miss,brake%20at%20the%20last%20second

Macpherson, T., Churchland, A., Sejnowski, T., DiCarlo, J., Kamitani, Y., Takahashi, H., & Hikida, T. (2021). Natural and Artificial Intelligence: A brief introduction to the interplay between AI and neuroscience research. *Neural Networks, 144*, 603-613.

Maguire, E. A., Gadian, D. G., Johnsrude, I. S., Good, C. D., Ashburner, J., Frackowiak, R. S. J., & Frith, C. D. (2000). Navigation-related structural change in the hippocampi of taxi drivers. *Proceedings of the National Academy of Sciences, 97*(8), 4398-4403.

McCausland, P. (2019). Self-driving Uber car that hit and killed woman did not recognize that pedestrians jaywalk. Retrieved from https://www.nbcnews.com/tech/tech-news/self-driving-uber-car-hit-killed-woman-did-not-recognize-n1079281

McClelland, J. L., McNaughton, B. L., & Lampinen, A. K. (2020). Integration of new information in memory: new insights from a complementary learning systems perspective. *Philosophical Transactions of the Royal Society B: Biological Sciences, 375*(1799), 20190637.

Millidge, B., Salvatori, T., Song, Y., Lukasiewicz, T., & Bogacz, R. (2022). Universal Hopfield Networks: A General Framework for Single-Shot Associative Memory Models. *Proc Mach Learn Res, 162*, 15561-15583.

MNE. (2025). Algorithms and other implementation details. Retrieved from https://mne.tools/stable/documentation/implementation.html#minimum-norm-estimates

Müller, M. (2007). Dynamic time warping. *Information Retrieval for Music and Motion, 2*, 69-84.

Munia, M. S., Hosseini, S. M., Nourani, M., Harvey, J., Dave, H., & Soc, I. C. (2021, Aug 09-12). *Imbalanced EEG Analysis Using One-Shot Learning with Siamese Neural Network.* Paper presented at the 9th IEEE International Conference on Healthcare Informatics (IEEE ICHI), Electr Network.

Murray, M. M., Brunet, D., & Michel, C. M. (2008). Topographic ERP analyses: a step-by-step tutorial review. *Brain Topography, 20*(4), 249-264.

Naser, M., & Bhattacharya, S. (2023). Towards Practical BCI-Driven Wheelchairs: A Systematic Review Study. *Ieee Transactions on Neural Systems and Rehabilitation Engineering, PP*, 1-1.

Navarro Cebrian, A., Knight, R., & Kayser, A. (2013). Error-Monitoring and Post-Error Compensations: Dissociation between Perceptual Failures and Motor Errors with and without Awareness. *The Journal of neuroscience : the official journal of the Society for Neuroscience, 33*, 12375-12383.

Neuper, C., Scherer, R., Reiner, M., & Pfurtscheller, G. (2005). Imagery of motor actions: Differential effects of kinesthetic and visual–motor mode of imagery in single-trial EEG. *Cognitive Brain Research, 25*(3), 668-677.
46